\documentclass[iicol,sn-mathphys-num]{sn-jnl}

\usepackage{graphicx}%
\usepackage{multirow}%
\usepackage{amsmath,amssymb,amsfonts}%
\usepackage{amsthm}%
\usepackage{mathrsfs}%
\usepackage[title]{appendix}%
\usepackage{xcolor}%
\usepackage{textcomp}%
\usepackage{manyfoot}%
\usepackage{booktabs}%
\usepackage{algorithm}%
\usepackage{algorithmicx}%
\usepackage{algpseudocode}%
\usepackage{listings}%
\usepackage{array}%
\usepackage{hyperref}%

\usepackage{bbm}
\usepackage{makecell}
\usepackage{threeparttable}
\usepackage{xspace}

\def\ie{\emph{i.e.}\@\xspace}

\theoremstyle{thmstyleone}%
\theoremstyle{thmstyletwo}%

\theoremstyle{thmstylethree}%
\begin{document}

\title[Article Title]{Day-Night Adaptation: An Innovative Source-free Adaptation Framework for Medical Image Segmentation}


\author[1]{\fnm{Ziyang} \sur{Chen}}\email{zychen@mail.nwpu.edu.cn}

\author[1]{\fnm{Yiwen} \sur{Ye}}\email{ywye@mail.nwpu.edu.cn}

\author*[1]{\fnm{Yongsheng} \sur{Pan}}\email{yspan@nwpu.edu.cn}

\author[2]{\fnm{Jingfeng} \sur{Zhang}}\email{jingfengzhang73@zju.edu.cn}

\author[1]{\fnm{Yanning} \sur{Zhang}}\email{ynzhang@nwpu.edu.cn}

\author*[1]{\fnm{Yong} \sur{Xia}}\email{yxia@nwpu.edu.cn}

\affil[1]{\orgdiv{National Engineering Laboratory for Integrated Aero-Space-Ground-Ocean Big Data Application Technology}, \orgname{Northwestern Polytechnical University}, \orgaddress{\city{Xi'an} \postcode{710072}, \country{China}}}



\affil[2]{\orgname{Ningbo No.2 Hospital}, \orgaddress{\city{Ningbo} \postcode{315000}, \country{China}}}


\abstract{Distribution shifts widely exist in medical images acquired from different medical centres, hindering the deployment of semantic segmentation models trained on one centre (source domain) to another (target domain). While unsupervised domain adaptation has shown significant promise in mitigating these shifts, it poses privacy risks due to sharing data between centres. To facilitate adaptation while preserving data privacy, source-free domain adaptation (SFDA) and test-time adaptation (TTA) have emerged as effective paradigms, relying solely on target domain data. 
However, SFDA requires a pre-collected target domain dataset before deployment.
TTA insufficiently exploit the potential value of test data, as it processes the test data only once.
Considering that most medical centres operate during the day and remain inactive at night in clinical practice, we propose a novel adaptation framework called \textbf{D}a\textbf{y}-\textbf{N}ight \textbf{A}daptation (DyNA) with above insights, which performs adaptation through day-night cycles without requiring access to source data. 
During the day, a low-frequency prompt is trained to adapt the frozen model to each test sample. We construct a memory bank for prompt initialization and develop a warm-up mechanism to enhance prompt training.
During the night, we reuse test data collected from the day and introduce a global student model to bridge the knowledge between teacher and student models, facilitating model fine-tuning while ensuring training stability.
Extensive experiments demonstrate that our DyNA outperforms existing TTA and SFDA methods on two benchmark medical image segmentation tasks.
Code will be available after the paper is published.}




\maketitle

\section{Introduction}\label{sec1}
Despite the significant advances in deep learning for medical image segmentation~\cite{seg_survey1}, deploying pre-trained segmentation models in real-world clinical applications remains challenging due to distribution shifts~\cite{domain_shift_1}. Such shifts, often resulting from variations in imaging protocols, operators, and scanners, are prevalent in medical images collected across different centers.
Unsupervised domain adaptation (UDA) methods~\cite{Parauda,ODADA,DoCR,UDA_Disentangle,BEAL,DALN} aim to address these shifts by aligning distributions between an unlabeled target domain and a labeled source domain. However, sharing data across medical centers is often restricted due to privacy concerns.

A more practical approach to protecting data privacy involves adapting the pre-trained source model to the target domain without accessing the source data. In this context, source-free domain adaptation (SFDA)~\cite{SFDA_survey1} and test-time adaptation (TTA)~\cite{TTA_survey,TENT} have emerged as promising solutions. Under the SFDA setting, a segmentation model pre-trained on labeled source data is distributed to a medical center, where it is fine-tuned using the center's unlabeled target data. In contrast, TTA adapts the source model exclusively with test data during the inference process, requiring fewer resources.
Despite their potential, both SFDA and TTA have limitations in practical applications. SFDA relies on a pre-collected target domain dataset, which is often unavailable. TTA typically processes each test sample only once, failing to fully exploit the value of the data. However, since medical centers often store test data for future use~\cite{storedata1,storedata2}, there is an opportunity to collect and reuse this data for further adaptation.

\begin{figure}[t]
  \centering
  \includegraphics[width=\columnwidth]{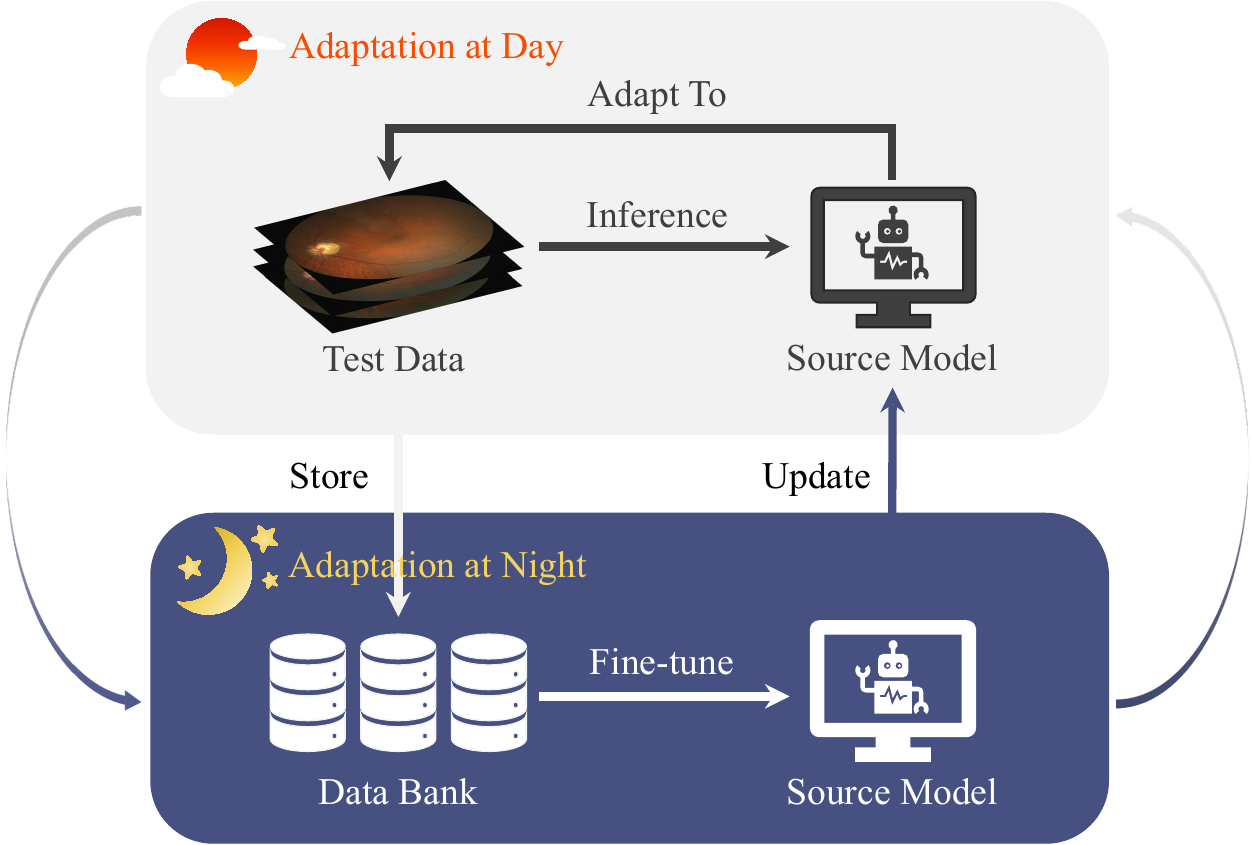}
   \caption{Pipeline of our DyNA framework. We continuously perform adaptation through day-night cycles. During the day, the pre-trained source model performs inference and adaptation on the test data, and the test data is stored in a data bank. During the night, the model's adaptation ability is further improved using the data bank.} 
\label{fig: pipeline}
\end{figure}

Based on these insights, we propose a novel source-free domain adaptation framework, called \textbf{D}a\textbf{y}-\textbf{N}ight \textbf{A}daptation (DyNA), designed to address real-world adaptation challenges. Inspired by the human brain, which processes information during the day and organizes it at night, DyNA adapts the source model to each test sample in a TTA manner during the day. These processed test samples are then collected and used to fine-tune the model at night. By alternating between day and night phases, DyNA continually adapts the pre-trained source model to the target domain.
However, two key challenges arise from the distinct nature of the day and night phases: (1) daytime adaptation may suffer from error accumulation due to domain shifts and limited supervision, and (2) nighttime adaptation, where only limited and unlabeled target data are available, with no access to source data, can result in unstable training. We address both challenges as follows.
During the day, we freeze the pre-trained source model and train a prompt for each test sample to adapt it from the target domain to the source domain, thereby preventing error accumulation from model updates. Specifically, we design a lightweight, low-frequency prompt that modifies only the low-frequency components of the test sample. A memory bank stores previous prompts, providing suitable initializations for new prompts. We train the prompt by aligning statistics in the batch normalization layers, replacing the source statistics with generated warm-up statistics to mitigate distribution shifts at the start of training.
At night, we introduce a global student model to enhance training stability. We freeze the trained prompt for each test sample and use the corresponding predictions as pseudo-labels. The source model, also frozen during the day, serves as the initialization for adaptation at night. We then fine-tune the source model using an enhanced self-training approach based on teacher-student models, with the global student model bridging the knowledge gap between the student and teacher. After fine-tuning, the teacher model updates the source model for the next day's deployment.
We evaluate DyNA on two benchmark medical image segmentation tasks: polyp segmentation and optic disc (OD) and optic cup (OC) segmentation. Our results demonstrate that DyNA outperforms all competing methods on both tasks.

The key contributions of this work are three-fold.
\begin{itemize}
    \item We develop DyNA, a novel framework for domain adaptation that continuously adapts a pre-trained source model to the target domain through day-night cycles without accessing the source data.
    \item We address the adaptation challenges by training a low-frequency prompt for each test sample to prevent error accumulation during the day and by developing a self-training approach that uses a global student model to enhance training stability during the night.
    \item We evaluate DyNA through extensive medical image segmentation experiments, demonstrating its effectiveness compared to existing TTA and SFDA methods.
\end{itemize}

A preliminary version of this work was presented at CVPR 2024~\cite{VPTTA}. In this paper, we have substantially revised and extended the original work. The main extensions include: (1) extending the Visual Prompt-based Test-Time Adaptation (VPTTA) method to the DyNA framework by adding the nighttime adaptation phase, which enables the reuse of test data for fine-tuning the source model; (2) introducing a global student model to improve training stability during the night; and (3) comparing DyNA with both existing TTA and SFDA methods.

\section{Related Work}
\label{sec:RelatedWorks}
\subsection{Unsupervised Domain Adaptation (UDA)}
UDA focuses on transferring knowledge from a source domain with labeled data to a target domain with unlabeled data. UDA methods can be broadly classified into three categories: input-level domain alignment, feature-level domain alignment, and output-level domain alignment.
Input-level domain alignment methods transform source-domain images to match the target-domain style while preserving original content. The image-translation techniques in this category include Cycle-GAN~\cite{cyclegan} and its variants~\cite{CYCMIS,CXDaGAN,Parauda}, as well as frequency-domain-based approaches~\cite{FDA,CAFT}. 
Wang~\emph{et al.}~\cite{CYCMIS} introduced semantic consistency constraints in image translation to improve segmentation consistency by leveraging both inter-domain and intra-domain consistency. 
Yang~\emph{et al.}~\cite{FDA} proposed a Fourier domain adaptation method that reduces distribution discrepancy by exchanging the low-frequency components between source-domain and target-domain images.
Feature-level domain alignment~\cite{gaussian_fa,ODADA,UDA_Disentangle} focuses on aligning the data distributions of the source and target domains in the feature space to capture domain-invariant features. Sun~\emph{et al.}~\cite{ODADA} introduced an orthogonal constraint to decompose domain-invariant and domain-specific features, using two independent domain discriminators for domain alignment. Shin~\emph{et al.}~\cite{UDA_Disentangle} used two independent encoders to disentangle intensity and non-intensity features, adapting the model to the more domain-invariant non-intensity features.
Output-level domain alignment involves extracting domain-invariant features through adversarial learning in the output space~\cite{ECSD,BEAL,DALN}. 
Wang~\emph{et al.}~\cite{BEAL} utilized two patch-based domain discriminators to improve the model's ability to produce low-entropy segmentation predictions and precise boundary predictions. 
Chen~\emph{et al.}~\cite{DALN} proposed a Nuclear-norm Wasserstein discrepancy, integrated with the task classifier, to serve as a domain discriminator for adversarial learning.
In contrast to these UDA approaches, we focus on addressing distribution shifts without accessing the source data, thereby protecting data privacy. 

\subsection{Test-time Adaptation (TTA)}
TTA aims to adapt a pre-trained source model to the test data during inference in a source-free and online manner~\cite{TENT}. The primary TTA methods involve updating the model by constructing self-supervised auxiliary tasks that guide the model to adapt to the test data~\cite{TENT,TTT,SAR,DLTTA,DomainAdaptor}.
Wang \emph{et al.}~\cite{TENT} proposed a test-time entropy minimization scheme that reduces generalization error by performing entropy minimization on the test data. 
Niu \emph{et al.}~\cite{SAR} removed partial noisy supervision with large uncertainty and encouraged model weights toward a flat minimum by sharpness-aware minimization.
Yang \emph{et al.}~\cite{DLTTA} developed a memory bank that adjusts the learning rate dynamically by computing the cosine similarity between current features and previous ones for each test image. 
Zhang \emph{et al.}~\cite{DomainAdaptor} proposed a generalized entropy minimization loss to enhance the utilization of test data information. 
While these methods perform well with reliable supervised information, they may struggle with error accumulation when the quality of supervision deteriorates.
Other approaches address distribution shifts by modifying the statistics in batch normalization (BN) layers~\cite{Dynamically,BN,DUA}. Wang \emph{et al.}~\cite{Dynamically} adapted each image dynamically through statistics-based distribution adaptation and prototype-based semantic adaptation. Mirza \emph{et al.}~\cite{DUA} continuously adjusted BN statistics to modify model feature representations. Although these methods minimize excessive model changes, they offer limited adaptation performance.
Despite the promise of TTA as a cost-effective approach, the utilization of test data in TTA is less sufficient.
In this paper, we address these by presenting a novel framework that performs adaptation through day-night cycles, enabling the reuse of test data collected from the day for fine-tuning during the night.

\subsection{Source-Free Domain Adaptation (SFDA)}
SFDA addresses distribution shifts by fine-tuning a pre-trained source model using only unlabeled target-domain data. Current SFDA methods are generally categorized into two types: data-centric and self-training.
Data-centric methods extended the UDA techniques to the SFDA scenario by either reconstructing a virtual source domain~\cite{VDM_DA,FSM,li2020model} or dividing source-like parts from the target domain~\cite{xia2021adaptive,chu2022denoised} to indirectly capture the distributions of source domain. Yang \emph{et al.}~\cite{FSM} introduced a Fourier-style-mining generator to create source-like images, which enhances domain alignment, and proposed a contrastive domain distillation module to narrow down the domain gap. 
Xia \emph{et al.}~\cite{xia2021adaptive} developed a contrastive category-wise matching module that leverages positive relations between target image pairs to enforce compactness within each category’s subspace.
Self-training methods, inspired from semi-supervised learning~\cite{saito2019semi,li2018semi}, focus on optimizing models using pseudo-labels~\cite{UPL-SFDA,PLPB} or employing a teacher-student paradigm~\cite{zhang2021source,PETS}. 
Wu \emph{et al.}~\cite{UPL-SFDA} used a duplicated pre-trained decoder with perturbations to increase prediction diversity for more reliable pseudo-labels and performed adaptation based on these labels. 
Li \emph{et al.}~\cite{PLPB} generated pseudo-labels from unlabeled target data, combining self-training with entropy minimization without requiring source data. 
Liu \emph{et al.}~\cite{PETS} periodically exchanged the weights between the static teacher model and the student model to integrate knowledge from past periods to reduce error accumulation.
Despite these advancements, the demand for pre-collected target domain dataset in existing SFDA methods may not always be met. 
In this paper, we present a novel framework that relies only on test data and enables continuous model adaptation during the inference process.

\begin{figure*}[!t]
  \centering
  \includegraphics[width=\textwidth]{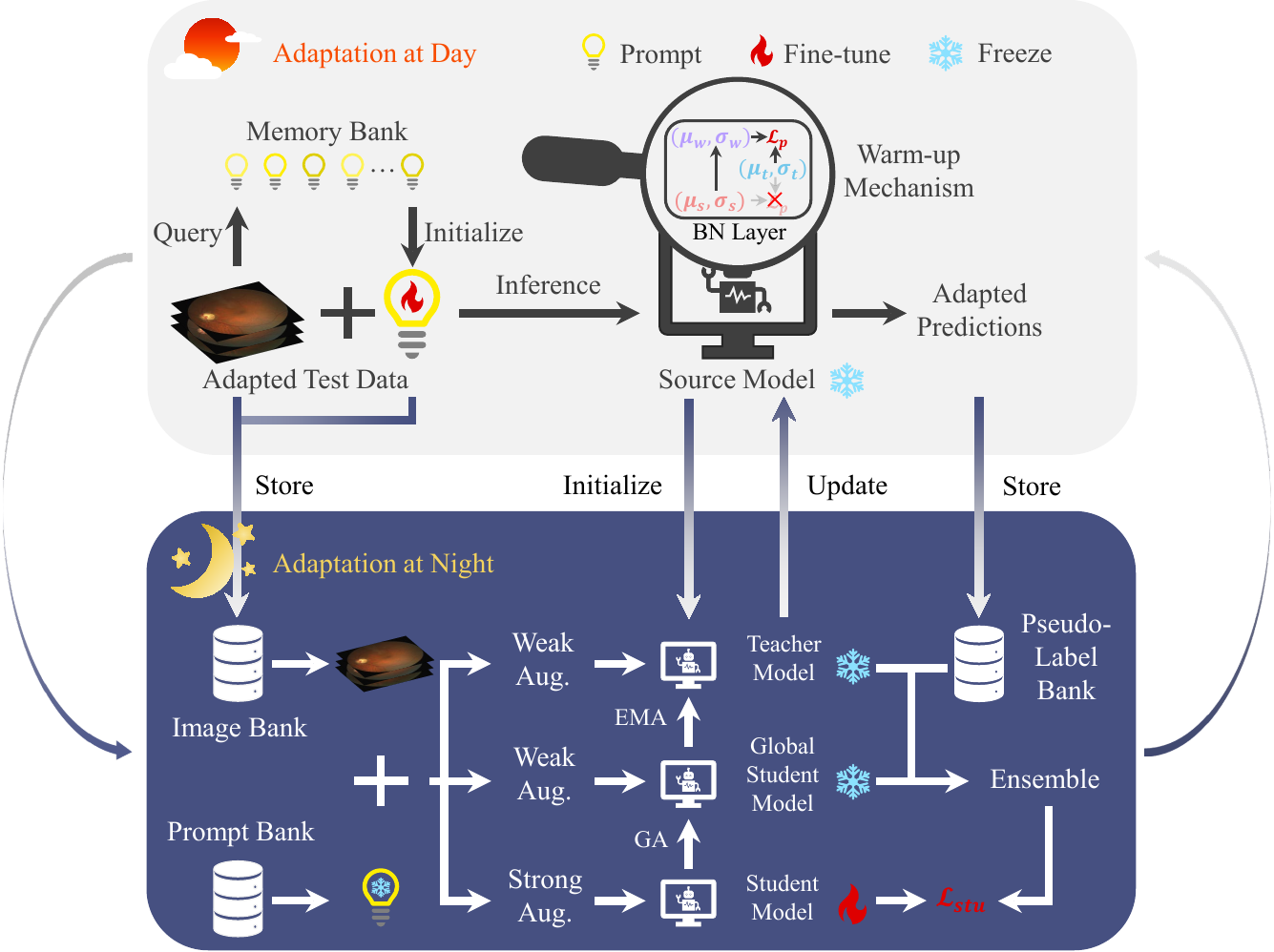}
   \caption{Overview of the adaptation approaches utilized during day and night in our DyNA framework: 
   (1) During the day, we freeze the pre-trained source model and train a learnable prompt for each test sample to boost adaptation performance. The test data, trained prompts, and adapted predictions are collected for use in the adaptation at night. 
   $\mu_s/\sigma_s$, $\mu_t/\sigma_t$, and $\mu_w/\sigma_w$ denote the source, test, and warm-up statistics (mean/standard deviation), respectively.
   (2) During the night, we freeze the trained prompts and fine-tune the source model using self-training on the collected data. A global student model is introduced to alleviate training instability. We integrate the pseudo-label and predictions of the global student model and teacher model to train the student model. The teacher model is updated by using the exponential moving average strategy and is subsequently frozen for inference the next day. 
   `BN': Abbreviation of `Batch Normalization'.
   `Aug.': Abbreviation of `Augmentation'. `GA': Abbreviation of `Global Average'. `EMA': Abbreviation of `Exponential Moving Average'.
   } 
\label{fig: overview}
\end{figure*}

\section{Method}
\label{sec:Method}

\subsection{Problem Definition and Method Overview}
Let $\mathbb{D}^{s} = \left \{ \mathcal{X}_{i}^{s}, \mathcal{Y}_{i}^{s} \right \}_{i=1}^{N^s}$ denote the labeled source domain dataset and $\mathbb{D}^{t} = \left \{ \mathcal{X}_{i}^{t}\right \}_{i=1}^{N^t}$ represent the unlabeled target domain dataset, where $\mathcal{X}_i^{*} \in \mathbb{R}^{H \times W \times C}$ is the $i$-th image with $C$ channels and dimensions $H \times W$, and $\mathcal{Y}_i^{*}$ is its corresponding label. 
Our DyNA framework alternates between adaptation approaches during day and night. 
We use a day-night cycle as an example to illustrate this process.

During the day, we train a low-frequency prompt $\mathcal{P}_i \in \mathbb{R}^{(\beta \times H) \times (\beta \times W) \times C}$ by minimizing the statistics aligning loss~\cite{ProSFDA,VPTTA}, where $\beta \in (0,1)$ is a small coefficient that ensures $\mathcal{P}_i$ focuses on the low-frequency components.
This prompt is applied to its corresponding test image $\mathcal{X}_{i}^{t}$ to produce a modified image $\tilde{\mathcal{X}}_{i}^{t}$, which is better adapted to the model $f^0:\mathcal{X} \rightarrow \mathcal{Y}$ pre-trained on $\mathbb{D}^{s}$. 
After inference, the test image, its prompt, and its adapted prediction are stored in the image bank, prompt bank, and pseudo-label bank, respectively. For clarity, we denote this set as $\mathcal{S}=\{\mathcal{X}_{i}^{t}, \mathcal{P}_i, \hat{\mathcal{Y}}_i\}_{i=1}^{N^{test}}$, where $\hat{\mathcal{Y}}_i$ represents the pseudo-label and $N^{test}$ is the number of test images collected from daytime.

During the night, the goal is to fine-tune the pre-trained source model $f^0$ using $\mathcal{S}$ to better fit the data distributions of $\mathbb{D}^{t}$. We first initialize the student model $f^0_{stu}$, global student model $f^0_{glo}$, and teacher model $f^0_{tea}$ from $f^0$. The student model $f^0_{stu}$ is supervised by an ensemble of predictions generated by $f^0_{glo}$ and $f^0_{tea}$, alongside pseudo-labels. After the night, the teacher model $f^0_{tea}$ is used to update the source model for the next day, setting $f^1=f^0_{tea}$. Additionally, $\mathcal{S}$ is reset for each night to prevent overfitting from repeated fine-tuning on the same data. Fig.~\ref{fig: overview} illustrates the adaptation approaches employed during day and night in our DyNA framework. We now delve into the adaptation process for the first day-night cycle.


\subsection{Adaptation at Day}
\subsubsection{Low-Frequency Prompt}
Let $\mathcal{F}(\cdot)$ and $\mathcal{F}^{-1}(\cdot)$ denote the Fast Fourier Transform (FFT) and Inverse Fast Fourier Transform (IFFT) operations~\cite{FFTW}, respectively. The amplitude and phase components are represented by $\left|\mathcal{F}(\cdot)\right|$ and $\mathcal{F}_\varphi(\cdot)$, with the low-frequency component $\left|\mathcal{F}(\cdot)\right|_l$ located at the center of $\left|\mathcal{F}(\cdot)\right|$. To facilitate efficient training during inference, we design a learnable low-frequency prompt $\mathcal{P}_i$ and pad it with ones to expand its size to $H \times W$. The padded $\mathcal{P}_i$ is then used to generate the adapted image $\tilde{\mathcal{X}}_{i}^{t}$ as follows:
\begin{equation}
\begin{aligned}
    \tilde{\mathcal{X}}_{i}^{t}=\mathcal{F}^{-1}([PadOne(\mathcal{P}_i) \odot \left|\mathcal{F}(\mathcal{X}_i^t)\right|, \mathcal{F}_\varphi(\mathcal{X}_i^t)]),
\end{aligned}
\label{eq1}
\end{equation}
where $PadOne(\cdot)$ denotes the padding operator, and $\odot$ represents element-wise multiplication.

\subsubsection{Prompt Initialization}
To ensure efficient and effective training, it is crucial to initialize the learnable prompt for each test image appropriately. We construct a memory bank $B$ that stores $K$ pairs of keys and values $\left \{ q_k, v_k \right \}_{k=1}^K$, where the keys and values are the low-frequency components of previous test images and their trained prompts, respectively. For the current test image $\mathcal{X}_i^t$, we extract its low-frequency component $\left|\mathcal{F}(\cdot)\right|_l$ and compute the cosine similarity between this component and each key $q_k$ in the memory bank $B$ as follows:
\begin{equation}
Cos(\left|\mathcal{F}(\cdot)\right|_l, q_k) = \frac{\langle\left|\mathcal{F}(\cdot)\right|_l, q_k\rangle}{\|\left|\mathcal{F}(\cdot)\right|_l\|*\|q_k\|},
\label{eq2}
\end{equation}
where $\langle \cdot, \cdot \rangle$ denotes the dot product and $\| \cdot \|$ represents the Euclidean norm. We then rank the similarity scores and retrieve the values of the $M$ most relevant keys from $B$ to construct a support set $R_i=\left \{ q_m, v_m \right \}_{m=1}^M$. The current prompt is initialized as a weighted sum of these values:
\begin{equation}
\mathcal{P}_i = \sum_{m=1}^{M} \omega_m v_m,
\label{eq: prompt_init}
\end{equation}
where each weight $\omega_m$ is determined as
\begin{equation}
\omega_m=\frac{e^{Cos(\left|\mathcal{F}(\cdot)\right|_l, q_m)}}{\sum_{j=1}^{M} e^{Cos(\left|\mathcal{F}(\cdot)\right|_l, q_j)}}.
\end{equation}
The memory bank $B$ is updated by the trained $\mathcal{P}_i$ using the First In, First Out (FIFO) principle.

\subsubsection{Prompt Training}
Inspired by~\cite{BN,ProSFDA}, distribution shifts primarily arise from the discrepancies between batch normalization (BN) statistics computed from test data (\ie, $\mu_t$ and $\sigma_t$) and those stored in the source model (\ie, $\mu_s$ and $\sigma_s$), where $\mu$ and $\sigma$ represent the mean and standard deviation, respectively.
However, directly aligning the source and test statistics within BN layers poses challenges, particularly due to the initially empty memory bank and ongoing distribution shifts at the start of the inference phase. To address this, we implement a warm-up mechanism that simulates the warm-up statistics (\ie, $\mu_{w}$ and $\sigma_{w}$) by combining the source and test statistics as follows:
\begin{equation}
\mu_{w} = \lambda \mu_t + (1 - \lambda) \mu_s, \,
\sigma_{w} = \lambda \sigma_t + (1 - \lambda) \sigma_s, 
\label{eq_mu_sig}
\end{equation}
where $\lambda = \frac{1}{\sqrt{i}/\tau +1}$, $i$ is the index of the current test image starting from 1, and $\tau$ is a temperature coefficient controlling the transition rate from warm-up to source statistics. The warm-up statistics $\mu_{w}$ and $\sigma_{w}$ are used for normalization instead of $\mu_s$ and $\sigma_s$. The prompt is trained by minimizing the statistics alignment loss~\cite{ProSFDA,VPTTA} in the BN layers:
\begin{equation}
\mathcal{L}_p=\frac{1}{h}\sum|\mu_{w}^h-\mu_t^h|+|{\sigma}_{w}^h-{\sigma}_t^h|,
\label{prompt_loss}
\end{equation}
where $h$ denotes the $h$-th BN layer. As inference progresses, the warm-up statistics gradually transition towards the source statistics, resulting in learning prompts from easy to difficult.

\subsubsection{Inference}
After training the prompt $\mathcal{P}_i$ by minimizing the statistics alignment loss given in Eq.~(\ref{prompt_loss}), the adapted image $\tilde{\mathcal{X}}_{i}^{t}$ is produced using trained $\mathcal{P}_i$ for the test image $\mathcal{X}_i^t$. The model then performs inference on $\tilde{\mathcal{X}}_{i}^{t}$ to obtain the adapted output $O_i$, calculated as $O_i=f^0(\tilde{\mathcal{X}}_{i}^{t})$, where $O_i$ also serves as the pseudo-label $\hat{\mathcal{Y}}_i$ during the night.

\subsection{Adaptation at Night}
\subsubsection{Self-training} 
We initialize the student model, global student model, and teacher model with the source model $f^0$ by $f^0_{stu} = f^0_{glo} = f^0_{tea} = f^0$. 
For each image in $\mathcal{S}$, we first generate the adapted image using the corresponding prompt, as outlined in Eq.~(\ref{eq1}). We then feed the strong augmentation variant into the student model and the weak augmentation variant into the global student model and teacher model. 
The weak augmentation strategy includes identity mapping, horizontal flipping, vertical flipping, and rotations of $90^\circ$, $180^\circ$, and $270^\circ$. Conversely, the strong augmentation strategy consists of brightness adjustment, contrast adjustment, Gamma transformation, Gaussian noise, and Gaussian blur.

To reduce noise in the supervision information, we combine the pseudo-label $\hat{\mathcal{Y}}$ with the predictions from the global student model and the teacher model (\ie, $\hat{\mathcal{P}}^{g}$ and $\hat{\mathcal{P}}^{t}$) as follows:
\begin{align}
\mathbb{I} = & \mathbbm{1} ((\hat{\mathcal{Y}}>T) \land (\hat{\mathcal{P}}^{g}>T) \land (\hat{\mathcal{P}}^{t}>T)) + \\ \nonumber
    & \mathbbm{1} ((\hat{\mathcal{Y}}<=T) \land (\hat{\mathcal{P}}^{g}<=T) \land (\hat{\mathcal{P}}^{t}<=T))
\end{align}
where $\mathbbm{1}$ denotes the indicator function, $\land$ represents the logical `and' operation, and $T$ is the threshold value. 

After filtering out unreliable supervision information, the training loss for the student model $f^0_{stu}$ is defined as:
\begin{align}
\mathcal{L}_{stu} = & \mathcal{L}_{bce}(\mathbb{I}*\hat{\mathcal{P}}^{s}, \label{student_loss}
\mathbb{I}*\hat{\mathcal{P}}^{g}) + \mathcal{L}_{bce}(\mathbb{I}*\hat{\mathcal{P}}^{s}, \mathbb{I}*\hat{\mathcal{P}}^{t}) + \\ \nonumber
& \mathcal{L}_{bce}(\mathbb{I}*\hat{\mathcal{P}}^{s}, \mathbb{I}*\hat{\mathcal{Y}}),
\end{align}
where $\hat{\mathcal{P}}^{s}$ denotes the prediction of the student model, and $\mathcal{L}_{bce}$ represents the binary cross-entropy loss.

\subsubsection{Model Update}
In each iteration, after updating the student model using $\mathcal{L}_{stu}$, the global student model is updated with the following global average strategy:
\begin{equation}
\label{global_update}
f^0_{glo} = \frac{r*f^0_{glo} + f^0_{stu}}{r+1},
\end{equation}
where $r$ denotes the iteration index, starting from 1. The teacher model is updated using the global student model according to the exponential moving average strategy:
\begin{equation}
\label{teacher_update}
f^0_{tea} = \alpha*f^0_{tea} + (1 - \alpha)*f^0_{glo},
\end{equation}
where $\alpha$ is a hyperparameter controlling the update rate. The fine-tuned teacher model is then used to update the source model for the next adaptation loop, setting $f^1 = f^0_{tea}$.

\section{Experiments and Results}
\subsection{Datasets}
\noindent \textbf{OD/OC Segmentation Dataset.} We constructed the OD/OC segmentation dataset using five public datasets collected from various medical centers, denoted as Domain A (RIM-ONE-r3~\cite{RIM_ONE_r3}), Domain B (REFUGE~\cite{REFUGE}), Domain C (ORIGA~\cite{ORIGA}), Domain D (REFUGE-Validation/Test~\cite{REFUGE}), and Domain E (Drishti-GS~\cite{Drishti-GS}). These domains contain 159, 400, 650, 800, and 101 images, respectively. Following the procedure outlined in~\cite{ProSFDA}, we cropped a region of interest (ROI) centered at OD with a size of $800\times 800$ for each image. Each ROI was then resized to $512\times 512$ and normalized using min-max normalization. Dice score coefficient ($DSC$) was employed as the evaluation metric for this task.

\noindent \textbf{Polyp Segmentation Dataset.} The polyp segmentation dataset is constructed by incorporating four public datasets obtained from different medical centers, denoted as Domain A (BKAI-IGH-NeoPolyp~\cite{BKAI_IGH_NeoPolyp}), Domain B (CVC-ClinicDB~\cite{CVC_ClinicDB}), Domain C (ETIS-LaribPolypDB~\cite{ETIS_LaribPolypDB}), and Domain D (Kvasir-Seg~\cite{Kvasir_Seg}). These datasets consist of 1000, 612, 196, and 1000 images, respectively. Following~\cite{PraNet}, we resized each image to $352\times 352$ and normalized the resized images using the statistics computed from the ImageNet dataset. The evaluation metrics for this task included $DSC$, the enhanced alignment metric ($E^{max}_\phi$)~\cite{Metric_E_max}, and structural similarity metric ($S_\beta$)~\cite{Metric_S_alpha}.

\noindent \textbf{Data Split.} During training the source model on each domain, all available data were utilized. For the application of TTA and SFDA methods, each domain was split in an 80:20 ratio, allocating 80\% of the data for inference and reserving 20\% for SFDA methods to perform fine-tuning, which was not accessed by TTA methods or our DyNA. The details are displayed in Table~\ref{tab:dataset}. To evaluate our DyNA approach across multiple day-night cycles in simulated deployment scenarios, the test data were divided into deployment periods of 10, 5, and 2 days, corresponding to 10\%, 20\%, and 50\% of the test data per cycle, respectively.


\begin{table}[!t]
  \centering
  \caption{Details of the data split used for OD/OC segmentation task and polyp segmentation task. The training data is exclusively available for SFDA methods.}
    \begin{tabular}{ccc}
    \toprule
    \multirow{2}{*}{Task} & \multirow{2}{*}{Domain} & Train (20\%)\\
    &&/Test (80\%)\\
    \midrule
    & A (RIM-ONE-r3) & 32/127\\
    \multirow{2}{*}{OD/}  & B (REFUGE) & 80/320\\
    \multirow{2}{*}{OC} & C (ORIGA) & 130/520\\
    & D (REFUGE-Validation/Test) & 160/640\\
    & E (Drishti-GS) & 20/81\\
    \midrule
    \multirow{4}{*}{Polyp} 
    & A (BKAI-IGH-NeoPolyp) & 200/800\\
    & B (CVC-ClinicDB) & 122/490\\
    & C (ETIS-LaribPolypDB) & 39/157\\
    & D (Kvasir-Seg) & 200/800\\
    \bottomrule
    \end{tabular}
  \label{tab:dataset}
\end{table}

\subsection{Implementation Details}
\label{subsec:implementation}
\noindent \textbf{Determination of Source and Target Domains.} For each task, we trained the source model on each individual domain (source domain) and tested it on the remaining domains (target domains) separately to compute mean metrics for evaluating the methods across various scenarios.
Specifically, we evaluate the methods across $4\times  5$ adaptation scenarios for the OD/OC segmentation task and $3\times 4$ adaptation scenarios for the polyp segmentation task.

\noindent \textbf{Segmentation Backbone.} Following~\cite{VPTTA}, we utilized a ResUNet~\cite{ProSFDA} based on ResNet-34~\cite{ResNet} as the segmentation backbone for the OD/OC segmentation task and a PraNet~\cite{PraNet} based on Res2Net~\cite{Res2Net} as the segmentation backbone for the polyp segmentation task.

\begin{table*}[!thb]
    \caption{Performance of our DyNA using different test data ratios, `Source Only' baseline, four competing TTA methods, and four competing SFDA methods on the OD/OC segmentation task. The best and second-best results in each column are highlighted in \textbf{bold} and \underline{underline}, respectively. 
    }
    \centering
    \resizebox{\textwidth}{!}{
    \begin{tabular}{cc|ccccc|c}
        \Xhline{1pt}
        \multicolumn{2}{c|}{\multirow{2}{*}{Methods}} & 
        \multicolumn{1}{c}{Domain A} & 
        \multicolumn{1}{c}{Domain B} & 
        \multicolumn{1}{c}{Domain C} & 
        \multicolumn{1}{c}{Domain D} & 
        \multicolumn{1}{c|}{Domain E} & 
        Average \\ 
        
        \Xcline{3-8}{0.4pt}
         & &
        $DSC$ &
        $DSC$ &
        $DSC$ &
        $DSC$ &
        $DSC$ &
        $DSC\uparrow$ \\
        \hline

         \multicolumn{2}{c|}{Source Only (ResUNet-34)} &
         $67.14$ & 	
         $75.75$ & 		
         $74.54$ & 	
         $52.19$ & 	
         $68.57$ & 		
         $67.64$ \\ 
         \hline


         \multicolumn{1}{c|}{\multirow{4}{*}{TTA}} & DLTTA~\cite{DLTTA} & 
         $74.32$ & 	
         $77.79$ & 		
         $75.87$ & 	
         $55.96$ & 	
         $71.26$ & 		
         $71.04$ \\ 

         \multicolumn{1}{c|}{} & DUA~\cite{DUA} &
         $72.15$ &
         $75.82$ &
         $74.91$ &
         $56.86$ &
         $72.48$ &
         $70.44$ \\ 

         \multicolumn{1}{c|}{} & SAR~\cite{SAR} & 
         $74.28$ & 	
         $77.23$ & 		
         $74.87$ & 	
         $58.65$ & 	
         $72.89$ & 		
         $71.58$ \\ 

         \multicolumn{1}{c|}{} & DomainAdaptor~\cite{DomainAdaptor} &
         $74.95$ & 	
         $76.54$ & 		
         $75.35$ & 	
         $56.79$ & 	
         $72.02$ & 		
         $71.13$ \\ 
         \hline

         \multicolumn{1}{c|}{\multirow{4}{*}{SFDA}} & FSM~\cite{FSM} &
         $74.14$ &
         $78.66$ &
         $77.29$ &
         $56.16$ &
         $72.51$ &
         $71.75$ \\ 

         \multicolumn{1}{c|}{} & UPL-SFDA~\cite{UPL-SFDA} & 
         $74.36$ & 	
         $76.39$ & 		
         $76.27$ & 	
         $\textbf{58.82}$ & 	
         $69.07$ & 		
         $70.98$ \\ 

         \multicolumn{1}{c|}{} & PETS~\cite{PETS} &
         $70.74$ &
         $75.77$ &
         $74.88$ &
         $57.23$ &
         $68.81$ &
         $69.49$  \\ 

         \multicolumn{1}{c|}{} & PLPB~\cite{PLPB} & 
         $72.72$ & 	
         $76.89$ & 		
         $77.14$ & 	
         $54.53$ & 	
         $74.74$ & 		
         $71.20$ \\ 
         \hline

         \multicolumn{1}{c|}{\multirow{3}{*}{DyNA (Ours)}} & $10\%$ &
         $74.86$ & 	
         $\underline{78.93}$ & 		
         $\underline{79.58}$ & 	
         $57.99$ & 	
         $78.33$ & 		
         $73.94$ \\ 

         \multicolumn{1}{c|}{} & $20\%$ &
         $\underline{75.36}$ & 	
         $\textbf{79.18}$ & 		
         $\textbf{79.77}$ & 	
         $\underline{58.29}$ & 	
         $\textbf{79.13}$ & 		
         $\textbf{74.34}$ \\ 

         \multicolumn{1}{c|}{} & $50\%$ &
         $\textbf{76.55}$ & 	
         $78.77$ & 		
         $78.94$ & 	
         $57.74$ & 	
         $\underline{78.57}$ & 		
         $\underline{74.12}$ \\ 
         
        \Xhline{1pt}
    \end{tabular}
    }
    \label{tab:Comparison_OD/OC}
\end{table*}

\begin{table*}[!htb]
    \caption{Performance of our DyNA using different test data ratios, `Source Only' baseline, and six competing methods on the polyp segmentation task. The best and second-best results in each column are highlighted in \textbf{bold} and \underline{underline}, respectively. 
    }
    \centering
    \resizebox{\textwidth}{!}{
    \begin{tabular}{cc|ccc ccc ccc ccc|ccc}
        \Xhline{1pt}
        \multicolumn{2}{c|}{\multirow{2}{*}{Methods}} & 
        \multicolumn{3}{c}{Domain A} & 
        \multicolumn{3}{c}{Domain B} & 
        \multicolumn{3}{c}{Domain C} & 
        \multicolumn{3}{c|}{Domain D} & 
        \multicolumn{3}{c}{Average} \\ 
        \Xcline{3-17}{0.4pt}
         & & 
        $DSC$ & $E^{max}_\phi$ & $S_\alpha$ &
        $DSC$ & $E^{max}_\phi$ & $S_\alpha$ &
        $DSC$ & $E^{max}_\phi$ & $S_\alpha$ &
        $DSC$ & $E^{max}_\phi$ & $S_\alpha$ &
        $DSC \uparrow$ & $E^{max}_\phi \uparrow$ & $S_\alpha \uparrow$ \\
        \hline

         \multicolumn{2}{c|}{Source Only (PraNet)} &
         $75.84$ & $85.46$ & $82.97$ & 
         $62.65$ & $78.09$ & $75.52$ & 
         $74.10$ & $84.63$ & $81.44$ & 
         $79.32$ & $88.10$ & $86.23$ & 
         $72.98$ & $84.07$ & $81.54$ \\
         \hline


         \multicolumn{1}{c|}{\multirow{4}{*}{TTA}} & DLTTA~\cite{DLTTA} & 
         $74.60$ & $85.45$ & $81.24$ & 
         $59.86$ & $72.34$ & $75.21$ & 
         $67.97$ & $80.38$ & $77.26$ & 
         $66.93$ & $77.54$ & $79.46$ & 
         $67.34$ & $78.93$ & $78.29$ \\

         \multicolumn{1}{c|}{} & DUA~\cite{DUA} & 
         $77.09$ & $86.94$ & $83.49$ & 
         $63.31$ & $76.94$ & $76.15$ & 
         $71.77$ & $82.18$ & $80.56$ & 
         $84.95$ & $92.50$ & $89.21$ & 
         $74.28$ & $84.64$ & $82.35$ \\

         \multicolumn{1}{c|}{} & SAR~\cite{SAR} & 
         $75.20$ & $86.04$ & $81.53$ & 
         $59.80$ & $72.37$ & $74.96$ & 
         $71.31$ & $82.72$ & $79.34$ & 
         $66.10$ & $76.79$ & $78.60$ & 
         $68.10$ & $79.48$ & $78.61$ \\ 

         \multicolumn{1}{c|}{} & DomainAdaptor~\cite{DomainAdaptor} & 
         $77.10$ & $87.24$ & $82.86$ & 
         $66.61$ & $79.14$ & $79.18$ & 
         $71.43$ & $82.31$ & $79.69$ & 
         $74.35$ & $83.93$ & $83.13$ & 
         $72.37$ & $83.16$ & $81.22$ \\
         \hline

         \multicolumn{1}{c|}{\multirow{4}{*}{SFDA}} & FSM~\cite{FSM} &
         $\underline{80.27}$ & $89.44$ & $\textbf{85.39}$ & 
         $66.47$ & $78.10$ & $76.54$ & 
         $\underline{77.71}$ & $\textbf{87.29}$ & $\textbf{83.66}$ & 
         $85.26$ & $93.15$ & $88.89$ & 
         $77.43$ & $87.00$ & $83.62$ \\ 

         \multicolumn{1}{c|}{} & UPL-SFDA~\cite{UPL-SFDA} & 
         $79.45$ & $88.26$ & $84.13$ & 
         $65.67$ & $76.94$ & $77.56$ & 
         $\textbf{77.76}$ & $\underline{86.71}$ & $\underline{83.62}$ & 
         $82.40$ & $91.37$ & $86.93$ & 
         $76.32$ & $85.82$ & $83.06$ \\ 

         \multicolumn{1}{c|}{} & PETS~\cite{PETS} & 
         $76.67$ & $86.01$ & $83.45$ & 
         $63.42$ & $78.95$ & $76.31$ & 
         $74.63$ & $85.25$ & $81.92$ & 
         $79.71$ & $88.13$ & $86.65$ & 
         $73.61$ & $84.59$ & $82.08$ \\ 

         \multicolumn{1}{c|}{} & PLPB~\cite{PLPB} & 
         $78.19$ & $88.11$ & $84.10$ & 
         $67.01$ & $80.61$ & $78.45$ & 
         $74.80$ & $85.00$ & $81.95$ & 
         $85.79$ & $93.61$ & $89.42$ & 
         $76.45$ & $86.83$ & $83.48$ \\
         \hline

         \multicolumn{1}{c|}{\multirow{3}{*}{DyNA (Ours)}} & $10\%$ &
         $79.92$ & $89.23$ & $84.83$ & 
         $73.86$ & $86.69$ & $82.73$ & 
         $76.41$ & $86.20$ & $82.81$ & 
         $86.63$ & $94.03$ & $90.06$ & 
         $79.20$ & $89.04$ & $85.11$ \\

         \multicolumn{1}{c|}{} & $20\%$ &
         $\textbf{80.46}$ & $\textbf{89.67}$ & $\underline{85.15}$ & 
         $\textbf{75.19}$ & $\textbf{87.93}$ & $\textbf{83.36}$ & 
         $76.67$ & $86.38$ & $82.91$ & 
         $\underline{86.76}$ & $\textbf{94.18}$ & $\underline{90.10}$ & 
         $\textbf{79.77}$ & $\textbf{89.54}$ & $\textbf{85.38}$ \\

         \multicolumn{1}{c|}{} & $50\%$ &
         $80.26$ & $\underline{89.52}$ & $85.02$ & 
         $\underline{74.89}$ & $\underline{87.69}$ & $\underline{83.26}$ & 
         $76.15$ & $86.03$ & $82.57$ & 
         $\textbf{86.79}$ & $\underline{94.07}$ & $\textbf{90.14}$ & 
         $\underline{79.52}$ & $\underline{89.33}$ & $\underline{85.25}$ \\
        \Xhline{1pt}
    \end{tabular}
    }
    \label{tab:Comparison_Polyp}
\end{table*}

\noindent \textbf{Hyper-parameters in Adaptation at Day.} 
During the day, we performed a one-iteration adaptation for each batch of test data, with a batch size of 1 in all experiments involving our DyNA. This setup was also applied to other competing TTA methods to ensure a fair comparison. 
We utilized the Adam optimizer with a learning rate of 0.05 for the OD/OC segmentation task and 0.01 for the polyp segmentation task. Following~\cite{VPTTA}, the hyper-parameters $\beta$ (size of the prompt), $K$ (size of the memory bank), $M$ (size of the support set), and $\tau$ (temperature coefficient during warm-up) were set to 0.01, 40, 16, and 5, respectively, for both segmentation tasks. Due to the complexity of PraNet's decoder, we only computed the prompt loss $\mathcal{L}_p$ in each BN layer of the encoder rather than the entire network for the polyp segmentation task.

\noindent \textbf{Hyper-parameters in Adaptation at Night.} 
During the night, we fine-tuned the source model for 10 epochs with a batch size of 4. We employed the SGD optimizer with a learning rate of 0.001 for the OD/OC segmentation task and 0.0003 for the polyp segmentation task. The hyper-parameter $T$ (threshold value) was empirically set to 0.5, and $\alpha$ (update rate for the teacher model) was set to 0.995 for the OD/OC segmentation task and 0.999 for the polyp segmentation task.

\subsection{Comparison with Other Methods}
We compared our DyNA with the `Source Only' baseline (training on the source domain without adaptation), four competing TTA methods, and four competing SFDA methods, as shown in Table~\ref{tab:Comparison_OD/OC} and Table~\ref{tab:Comparison_Polyp}. The competing TTA methods include an entropy-based method (SAR~\cite{SAR}), a method utilizing a dynamic learning rate (DLTTA~\cite{DLTTA}), a method combining entropy minimization and statistics fusion (DomainAdaptor~\cite{DomainAdaptor}), and a method that modifies the BN statistics (DUA~\cite{DUA}). The competing SFDA methods include a data-centric approach (FSM~\cite{FSM}) and three self-training methods (UPL-SFDA~\cite{UPL-SFDA}, PETS~\cite{PETS}, and PLPB~\cite{PLPB}), with PETS employing self-training based on a teacher-student paradigm. Specifically, we reproduced these methods using the same baseline as our DyNA for a fair comparison. 
The results displayed for each domain are calculated using that domain as the source domain while deploying the corresponding method on the remaining domains.

\subsubsection{OD/OC Segmentation Task}
Table~\ref{tab:Comparison_OD/OC} presents the comparison results for OD/OC segmentation. It is evident that SFDA and TTA methods achieve broadly similar performance, with all methods outperforming the “Source Only” baseline, thereby demonstrating the effectiveness of adaptation in addressing distribution shifts. Our DyNA, which implements adaptation across day-night cycles, outperforms both SFDA and TTA methods in most scenarios and achieves the best overall performance, highlighting the effectiveness of our day-night adaptation strategy. Additionally, our DyNA exhibits robustness across various test data ratios, and this can be attributed to that we maintain training stability during the night by introducing the global student model and freeze model parameters during the day to avoid error accumulation.

\subsubsection{Polyp Segmentation Task}
Table~\ref{tab:Comparison_Polyp} presents the comparison results for polyp segmentation. On this task, SFDA methods generally demonstrate superior performance compared to TTA methods, underscoring the importance of utilizing unlabeled target training data. Notably, some TTA methods incorporating entropy minimization (\ie, DLTTA, SAR, and DomainAdaptor) perform worse than the `Source Only’ baseline, which can be attributed to the inherent challenges associated with polyp segmentation, as polyps are often concealed and thus more difficult to detect amid distribution shifts. Our DyNA exhibits superior overall performance across both tasks, achieving the best results in almost all scenarios, thereby emphasizing its exceptional applicability and robustness. This suggests that unlabeled target training data may not be indispensable for achieving strong performance; rather, our approach leverages previous test data to facilitate effective model transfer to the target domain. Similar to the observations on the OD/OC segmentation task, our DyNA maintains strong adaptation performance across various test data ratios, remaining largely unaffected by catastrophic forgetting.

\begin{table*}[!htb]
    \caption{Results of ablation study on the OD/OC segmentation task. The best and second-best results in each column are highlighted in \textbf{bold} and \underline{underline}, respectively. `P': Abbreviation of prompt. `M': Abbreviation of memory bank. `W': Abbreviation of warm-up. $f_{tea}\leftarrow f_{*}$ means utilizing $f_{*}$ to update the teacher model. $f_{glo}$ indicates introducing the global student model to provide predictions for better supervision information.
    }
    \centering
    \resizebox{\textwidth}{!}{
    \begin{tabular}{ccc|ccc|c|ccccc|c}
        \Xhline{1pt}
        \multicolumn{7}{c|}{Methods} & 
        \multicolumn{1}{c}{\multirow{2}{*}{Do. A}} & 
        \multicolumn{1}{c}{\multirow{2}{*}{Do. B}} & 
        \multicolumn{1}{c}{\multirow{2}{*}{Do. C}} & 
        \multicolumn{1}{c}{\multirow{2}{*}{Do. D}} & 
        \multicolumn{1}{c|}{\multirow{2}{*}{Do. E}} & 
        \multirow{2}{*}{Average} \\ 
        \Xcline{1-7}{0.4pt}
        \multicolumn{3}{c|}{Day} & \multicolumn{3}{c|}{Night} & \multicolumn{1}{c|}{Test Data Ratio} & & & & &\\
        \hline
        P & M & W & $f_{glo}$ & $f_{tea}\leftarrow f_{glo}$ & $f_{tea}\leftarrow f_{stu}$ & $10\% / 20\% / 50\%$&
        $DSC$ & 
        $DSC$ & 
        $DSC$ & 
        $DSC$ & 
        $DSC$ & 
        $DSC\uparrow$ \\
        \hline

         \checkmark & & & & & & --- &
         $68.05$ & 
         $76.10$ & 
         $75.13$ & 
         $52.80$ & 
         $70.33$ & 
         $68.48$ \\

         \checkmark & \checkmark & \multicolumn{1}{c|}{} & & & & --- &
         $70.46$ & 
         $77.38$ & 
         $76.28$ & 
         $54.35$ & 
         $73.50$ & 
         $70.39$ \\

         \checkmark & & \multicolumn{1}{c|}{\checkmark} & & & & --- &
         $75.19$ & 
         $77.31$ & 
         $76.56$ & 
         $54.85$ & 
         $72.21$ & 
         $71.22$ \\

         \checkmark & \checkmark & \multicolumn{1}{c|}{\checkmark} & & & & --- &
         $74.27$ & 	
         $78.09$ & 		
         $76.87$ & 	
         $55.92$ & 	
         $75.93$ & 		
         $72.22$ \\ 
         \hline

         \checkmark & \checkmark & \multicolumn{1}{c|}{\checkmark} & & & \checkmark & $10\%$ &
         $73.43$ & 	
         $78.96$ & 	
         $76.65$ &	
         $41.82$ &	
         $73.48$ & 	
         $68.87$ \\

         \checkmark & \checkmark & \multicolumn{1}{c|}{\checkmark} & \checkmark & & \checkmark & $10\%$ &
         $74.33$ & 
         $79.08$ & 
         $78.88$ & 
         $56.75$ & 
         $75.88$ & 
         $72.98$ \\

         \checkmark & \checkmark & \multicolumn{1}{c|}{\checkmark} & \checkmark & \checkmark & & $10\%$ &
         $74.86$ & 	
         $78.93$ & 		
         $\underline{79.58}$ & 	
         $\underline{57.99}$ & 	
         $78.33$ & 		
         $73.94$ \\ 
         \hline

         \checkmark & \checkmark & \multicolumn{1}{c|}{\checkmark} & & & \checkmark & $20\%$ &
         $76.12$ & 	
         $78.53$ & 	
         $75.17$ & 	
         $53.49$ & 	
         $74.36$ & 	
         $71.53$ \\

         \checkmark & \checkmark & \multicolumn{1}{c|}{\checkmark} & \checkmark & & \checkmark & $20\%$ &
         $\underline{76.20}$ & 	
         $\underline{79.17}$ & 	
         $78.37$ & 	
         $57.34$ & 	
         $78.30$ & 	
         $73.88$ \\

         \checkmark & \checkmark & \multicolumn{1}{c|}{\checkmark} & \checkmark & \checkmark & & $20\%$ &
         $75.36$ & 	
         $\textbf{79.18}$ & 		
         $\textbf{79.77}$ & 	
         $\textbf{58.29}$ & 	
         $\textbf{79.13}$ & 		
         $\textbf{74.34}$ \\ 
         \hline

         \checkmark & \checkmark & \multicolumn{1}{c|}{\checkmark} & & & \checkmark & $50\%$ &
         $75.56$ & 	
         $78.30$ & 	
         $77.04$ & 	
         $56.88$ & 	
         $76.30$ & 	
         $72.82$ \\

         \checkmark & \checkmark & \multicolumn{1}{c|}{\checkmark} & \checkmark & & \checkmark & $50\%$ &
         $76.03$ & 
         $78.98$ & 
         $77.66$ & 
         $55.92$ & 
         $78.26$ & 
         $73.37$ \\	 	 

         \checkmark & \checkmark & \multicolumn{1}{c|}{\checkmark} & \checkmark & \checkmark & & $50\%$ &
         $\textbf{76.55}$ & 	
         $78.77$ & 		
         $78.94$ & 	
         $57.74$ & 	
         $\underline{78.57}$ & 		
         $\underline{74.12}$ \\ 
        \Xhline{1pt}
    \end{tabular}
    }
    \label{tab:ablation}
\end{table*}

\begin{figure*}[!tb]
    \centering
    \includegraphics[width=\textwidth]{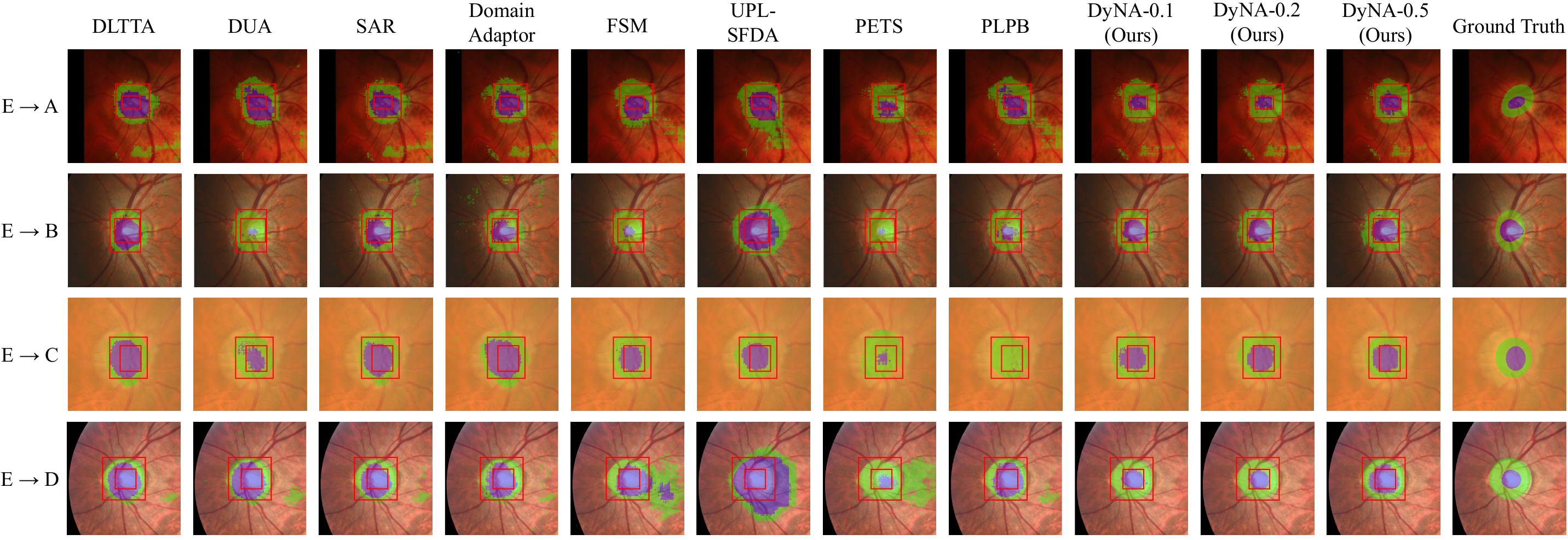}
    \caption{Visualization of OD/OC segmentation results obtained by four competing TTA methods, four competing SFDA methods, and our DyNA. From top to bottom: segmentation results and ground truth obtained from the scenario denoted as ``source $\rightarrow$ target''. From left to right: segmentation results of the TTA methods (DLTTA, DUA, SAR, and DomainAdaptor), SFDA methods (FSM, UPL-SFDA, PETS, and PLPB), and our DyNA using different test data ratios, followed by the ground truth. To highlight potential over- or under-segmentation, the ground truth bounding boxes of OD and OC are overlaid on each segmentation result. Best viewed in color.}
    \label{fig:visual-optic}
\end{figure*}

\subsection{Ablation Study}
To assess the contributions of our adaptation strategies for day-night cycles, we conducted ablation studies on the OD/OC segmentation task. For adaptation at day, we evaluated the contributions of three key components: the low-frequency prompt, memory bank-based initialization, and warm-up mechanism. Since adaptation at day is not influenced by variations in test data ratios, we only conducted experiments with different test data ratios for adaptation at night. Our results presented in Table~\ref{tab:ablation} indicate that (1) using only the low-frequency prompt yields limited performance gains due to the inadequate training of the prompt; (2) both the memory bank and warm-up mechanism enhance prompt training; and (3) optimal performance is achieved when all components are utilized together, highlighting their complementary effects. For adaptation at night, we analyzed the effectiveness of our proposed global student model and discussed which model is more suitable for updating the teacher model. The findings demonstrate that (1) introducing the global student model enhances performance; (2) updating the teacher model using the global student model yields superior results; and (3) the impact of the global student model is more pronounced in long-term adaptation scenarios (\ie, when the test data ratio is $10\%$).

\begin{figure*}[!tb]
    \centering
    \includegraphics[width=\textwidth]{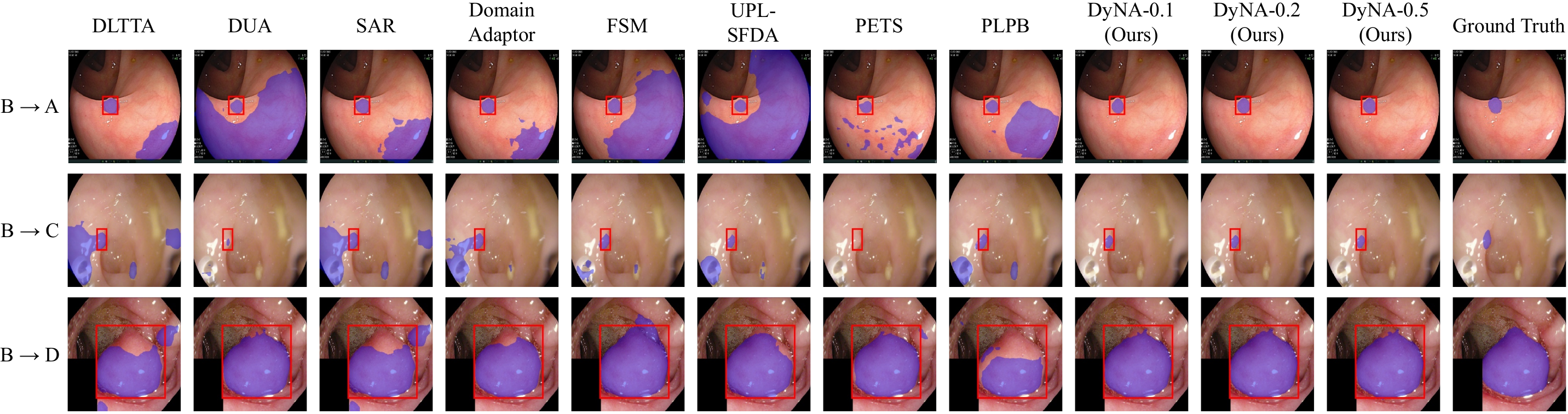}
    \caption{Visualization of polyp segmentation results obtained by four competing TTA methods, four competing SFDA methods, and our DyNA. From top to bottom: segmentation results and ground truth obtained from the scenario denoted as ``source $\rightarrow$ target''. From left to right: segmentation results of the TTA methods (DLTTA, DUA, SAR, and DomainAdaptor), SFDA methods (FSM, UPL-SFDA, PETS, and PLPB), and our DyNA using different test data ratios, followed by the ground truth. To highlight potential over- or under-segmentation, the ground truth bounding boxes of foreground are overlaid on each segmentation result. Best viewed in color.}
    \label{fig:visual-polyp}
\end{figure*}

\begin{table*}[!htb]
    \caption{Results of various combinations of our DyNA and other TTA/SFDA methods using different test data ratios on the OD/OC segmentation task. The \textcolor[rgb]{1,0,0}{red} numbers indicate the performance gain achieved by the combination over the method without our DyNA.}
    \centering
    \resizebox{\textwidth}{!}{
    \begin{tabular}{c|c|c|ccccc|c}
        \Xhline{1pt}
        \multicolumn{3}{c|}{Methods} & 
        \multicolumn{1}{c}{\multirow{2}{*}{Do. A}} & 
        \multicolumn{1}{c}{\multirow{2}{*}{Do. B}} & 
        \multicolumn{1}{c}{\multirow{2}{*}{Do. C}} & 
        \multicolumn{1}{c}{\multirow{2}{*}{Do. D}} & 
        \multicolumn{1}{c|}{\multirow{2}{*}{Do. E}} & 
        \multirow{2}{*}{Average} \\ 
        \Xcline{1-3}{0.4pt}
        \multicolumn{1}{c|}{\multirow{2}{*}{Day}} & \multicolumn{1}{c|}{\multirow{2}{*}{Night}} & \multicolumn{1}{c|}{Test Data Ratio} & & & & & \\
        \Xcline{3-9}{0.4pt}
         &  & $10\% / 20\% / 50\%$&
        $DSC$ & 
        $DSC$ & 
        $DSC$ & 
        $DSC$ & 
        $DSC$ & 
        $DSC\uparrow$ \\
        \hline

        \multicolumn{1}{c|}{\multirow{1}{*}{DLTTA~\cite{DLTTA}}} & \multicolumn{1}{c|}{\multirow{1}{*}{-}} & - &
        $74.32$ & 	
        $77.79$ & 		
        $75.87$ & 	
        $55.96$ & 	
        $71.26$ & 		
        $71.04$ \\
        \hline

        \multicolumn{1}{c|}{\multirow{3}{*}{DLTTA~\cite{DLTTA}}} & \multicolumn{1}{c|}{\multirow{3}{*}{DyNA (Ours)}} & $10\%$ &
        $76.98$ & 
        $78.64$ & 
        $78.11$ & 
        $59.76$ & 
        $71.78$ & 
        $73.05$ \textcolor[rgb]{1,0,0}{$+2.01$} \\ 

         & & $20\%$ &
        $76.54$ & 
        $78.79$ & 
        $77.66$ & 
        $58.63$ & 
        $72.10$ & 
        $72.74$ \textcolor[rgb]{1,0,0}{$+1.70$} \\

         & & $50\%$ &
        $75.30$ & 
        $78.13$ & 
        $76.89$ & 
        $56.77$ & 
        $71.71$ & 
        $71.76$ \textcolor[rgb]{1,0,0}{$+0.72$} \\ 
        \hline

        \multicolumn{1}{c|}{\multirow{1}{*}{SAR~\cite{SAR}}} & \multicolumn{1}{c|}{\multirow{1}{*}{-}} & - &
        $74.28$ & 	
        $77.23$ & 		
        $74.87$ & 	
        $58.65$ & 	
        $72.89$ & 		
        $71.58$ \\
        \hline

        \multicolumn{1}{c|}{\multirow{3}{*}{SAR~\cite{SAR}}} & \multicolumn{1}{c|}{\multirow{3}{*}{DyNA (Ours)}} & $10\%$ &
        $75.58$ & 
        $77.27$ & 
        $77.03$ & 
        $61.65$ & 
        $73.20$ & 
        $72.95$ \textcolor[rgb]{1,0,0}{$+1.37$} \\

         & & $20\%$ &
        $75.66$ & 
        $77.31$ & 
        $77.03$ & 
        $61.80$ & 
        $73.13$ & 
        $72.99$ \textcolor[rgb]{1,0,0}{$+1.41$} \\

         & & $50\%$ &
        $74.97$ & 
        $77.24$ & 
        $76.08$ & 
        $59.39$ & 
        $73.47$ & 
        $72.23$ \textcolor[rgb]{1,0,0}{$+0.65$} \\
        \hline

        \multicolumn{1}{c|}{\multirow{3}{*}{-}} & \multicolumn{1}{c|}{\multirow{3}{*}{FSM~\cite{FSM}}} & $10\%$ &
        $75.06$ & 
        $77.49$ & 
        $76.47$ & 
        $56.05$ & 
        $71.88$ & 
        $71.39$ \\

         & & $20\%$ &
        $74.02$ & 
        $77.25$ & 
        $76.29$ & 
        $55.37$ & 
        $71.86$ & 
        $70.96$ \\

         & & $50\%$ &
        $71.38$ & 
        $76.84$ & 
        $75.57$ & 
        $53.92$ & 
        $71.59$ & 
        $69.86$ \\
        \hline

        \multicolumn{1}{c|}{\multirow{3}{*}{DyNA (Ours)}} & \multicolumn{1}{c|}{\multirow{3}{*}{FSM~\cite{FSM}}} & $10\%$ &
        $75.27$ & 
        $78.29$ & 
        $76.62$ & 
        $57.47$ & 
        $75.81$ & 
        $72.69$ \textcolor[rgb]{1,0,0}{$+1.30$} \\

         & & $20\%$ &
        $75.14$ & 
        $78.23$ & 
        $76.61$ & 
        $57.13$ & 
        $75.36$ & 
        $72.49$ \textcolor[rgb]{1,0,0}{$+1.53$} \\

         & & $50\%$ &
        $74.77$ & 
        $78.20$ & 
        $76.71$ & 
        $56.53$ & 
        $75.17$ & 
        $72.28$ \textcolor[rgb]{1,0,0}{$+2.42$} \\
        \hline

        \multicolumn{1}{c|}{\multirow{3}{*}{-}} & \multicolumn{1}{c|}{\multirow{3}{*}{PLPB~\cite{PLPB}}} & $10\%$ &
        $71.54$ & 
        $77.29$ & 
        $75.06$ & 
        $53.21$ & 
        $70.59$ & 
        $69.54$ \\

         & & $20\%$ &
        $71.19$ & 
        $76.42$ & 
        $75.16$ & 
        $52.99$ & 
        $70.21$ & 
        $69.20$ \\

         & & $50\%$ &
        $69.50$ & 
        $75.81$ & 
        $74.82$ & 
        $51.62$ & 
        $67.80$ & 
        $67.91$ \\
        \hline

        \multicolumn{1}{c|}{\multirow{3}{*}{DyNA (Ours)}} & \multicolumn{1}{c|}{\multirow{3}{*}{PLPB~\cite{PLPB}}} & $10\%$ &
        $73.42$ & 
        $78.28$ & 
        $76.46$ & 
        $56.08$ & 
        $76.18$ & 
        $72.08$ \textcolor[rgb]{1,0,0}{$+2.54$} \\

         & & $20\%$ &
        $73.61$ & 
        $78.46$ & 
        $76.37$ & 
        $55.83$ & 
        $75.48$ & 
        $71.95$ \textcolor[rgb]{1,0,0}{$+2.75$} \\

         & & $50\%$ &
        $73.03$ & 
        $78.29$ & 
        $76.28$ & 
        $56.64$ & 
        $75.91$ & 
        $72.03$ \textcolor[rgb]{1,0,0}{$+4.12$} \\
        \Xhline{1pt}
    \end{tabular}
    }
    \label{tab:discuss_combine}
\end{table*}

\begin{table*}[!htb]
    \caption{Results of employing different models for the next day's deployment on the OD/OC segmentation task. The best and second-best results in each column are highlighted in \textbf{bold} and \underline{underline}, respectively.}
    \centering
    \begin{tabular}{c|c|ccccc|c}
        \Xhline{1pt}
        \multicolumn{2}{c|}{Methods} & 
        \multicolumn{1}{c}{\multirow{2}{*}{Do. A}} & 
        \multicolumn{1}{c}{\multirow{2}{*}{Do. B}} & 
        \multicolumn{1}{c}{\multirow{2}{*}{Do. C}} & 
        \multicolumn{1}{c}{\multirow{2}{*}{Do. D}} & 
        \multicolumn{1}{c|}{\multirow{2}{*}{Do. E}} & 
        \multirow{2}{*}{Average} \\ 
        \Xcline{1-2}{0.4pt}
        \multicolumn{1}{c|}{\multirow{2}{*}{Model}} & \multicolumn{1}{c|}{Test Data Ratio} & & & & & \\
        \Xcline{2-8}{0.4pt}
         & $10\% / 20\% / 50\%$&
        $DSC$ & 
        $DSC$ & 
        $DSC$ & 
        $DSC$ & 
        $DSC$ & 
        $DSC\uparrow$ \\
        \hline
        
        \multicolumn{1}{c|}{\multirow{3}{*}{$f_{stu}$}} & $10\%$ &
        $68.29$ &  
        $73.58$ &  
        $76.41$ &  
        $39.62$ &  
        $73.00$ &  
        $66.18$ \\ 
        
         & $20\%$ &
        $71.03$ &  
        $78.37$ &  
        $77.99$ &  
        $49.82$ &  
        $77.01$ &  
        $70.85$ \\ 
        
         & $50\%$ &
        $75.51$ &  
        $78.70$ &  
        $77.70$ &  
        $57.17$ &  
        $78.03$ &  
        $73.42$ \\ 
        \hline

        \multicolumn{1}{c|}{\multirow{3}{*}{$f_{glo}$}} & $10\%$ &
        $75.75$ &  
        $79.03$ &  
        $78.38$ &  
        $52.89$ &  
        $68.31$ &  
        $70.87$ \\  

         & $20\%$ &
        $75.23$ &  
        $\underline{79.15}$ &  
        $77.60$ &  
        $57.92$ &  
        $78.45$ &  
        $73.67$ \\ 

         & $50\%$ &
        $\underline{76.25}$ &  
        $78.80$ &  
        $78.37$ &  
        $56.88$ &  
        $77.80$ &  
        $73.62$ \\   
        \hline
        
        \multicolumn{1}{c|}{\multirow{3}{*}{$f_{tea}$}} & $10\%$ &
        $74.86$ & 
        $78.93$ & 
        $\underline{79.58}$ & 
        $\underline{57.99}$ & 
        $78.33$ & 
        $73.94$ \\ 

         & $20\%$ &
        $75.36$ & 
        $\textbf{79.18}$ & 
        $\textbf{79.77}$ & 
        $\textbf{58.29}$ & 
        $\textbf{79.13}$ & 
        $\textbf{74.34}$ \\

         & $50\%$ &
        $\textbf{76.55}$ & 
        $78.77$ & 
        $78.94$ & 
        $57.74$ & 
        $\underline{78.57}$ & 
        $\underline{74.12}$ \\ 
        \Xhline{1pt}
    \end{tabular}
    \label{tab:discuss_optimal}
\end{table*}

\section{Discussion}
\subsection{Visualization of Segmentation Results}
For qualitative analysis, we conducted two case studies: one using the source model trained on Domain E for the OD/OC segmentation task, and the other using the source model trained on Domain B for the polyp segmentation task. We visualized the ground truth and segmentation results of eight competing methods alongside our DyNA across different test data ratios in various scenarios denoted as `source $ \rightarrow $ target', as shown in Fig.~\ref{fig:visual-optic} and Fig.~\ref{fig:visual-polyp}. To facilitate comparison, we outlined the foreground regions of the ground truth with bounding boxes, which were then overlaid on each segmentation result. In Fig.~\ref{fig:visual-optic}, the OD and OC regions are highlighted in green and blue, respectively. The results reveal that our DyNA exhibits reduced over-segmentation and under-segmentation compared to other methods across all samples, highlighting the effectiveness of the proposed day-night adaptation strategy. In Fig.~\ref{fig:visual-polyp}, our DyNA shows superior segmentation performance for both large objects (row 3) and small objects (rows 1 and 2) compared to other competing methods.

\subsection{Combination of Our DyNA and Other Methods}
To explore the extensibility of our DyNA, we replaced the adaptation processes used during day and night with various TTA and SFDA methods and repeated experiments on the OD/OC segmentation task with different test data ratios. For comparison, we also conducted experiments under the same setup as our DyNA for two TTA methods (\ie, DLTTA and SAR) and two SFDA methods (\ie, FSM and PLPB). Since the test data ratio does not affect TTA methods, we inherited their results from Table~\ref{tab:Comparison_OD/OC}. The findings in Table~\ref{tab:discuss_combine} indicate that our DyNA enhances the performance of both TTA and SFDA methods across all test data ratios, showcasing its good extensibility. Unlike other SFDA methods that experience performance degradation over prolonged deployment, those equipped with our DyNA demonstrate stable and superior performance across all test data ratios, highlighting the robustness and effectiveness of our approach.

\subsection{Optimal Model for Next-Cycle Deployment}
Given the presence of three models (\ie, the student model, global student model, and teacher model) during the night, we conducted experiments on the OD/OC segmentation task to identify the optimal model for next-cycle inference. The results are summarized in Table~\ref{tab:discuss_optimal}. Our analysis reveals that (1) using either the student model or global student model for next-cycle inference may result in unexpected performance degradation, particularly when the test data ratio is set to 10\%; and (2) using the teacher model consistently achieves the best overall performance across all test data ratios.

\section{Conclusion}
In this paper, we propose a novel framework called DyNA, which continually adapts the pre-trained source model through day-night cycles to better align with the realistic scenarios and sufficiently utilize test data. 
During the day, we freeze the model parameters and train a lightweight low-frequency prompt for each test image while constructing a memory bank for better prompt initialization. We also implement a warm-up mechanism based on statistics fusion to facilitate the training process. 
During the night, we fine-tune the model using a self-training paradigm. To ensure stable fine-tuning, we introduce a global student model that integrates historical knowledge to update the teacher model using an exponential moving average strategy. 
Extensive experimental results conducted on two medical image segmentation tasks across multiple scenarios demonstrate the superiority of our DyNA over existing TTA and SFDA methods and confirm that DyNA can be combined with other methods to enhance their performance.

\noindent{\textbf{Acknowledgments.}} This work was supported in part by the National Natural Science Foundation of China (Nos. 62171377, 6240012686), and in part by the Natural Science Foundation of Ningbo City, China, under Grant 2021J052.

\noindent{\textbf{Data availability statements.}} 
The findings of this study are supported by the OD/OC segmentation dataset, which consists of five sub-datasets (i.e., RIM-ONE-r3, REFUGE, ORIGA, REFUGE-Validation/Test, and Drishti-GS), and the polyp segmentation dataset, which includes four sub-datasets (i.e., BKAI-IGH-NeoPolyp, CVC-ClinicDB, ETIS-LaribPolypDB, and Kvasir-Seg).
The OD/OC segmentation dataset and the polyp segmentation dataset are openly available at \href{https://drive.google.com/drive/folders/1axgu3-65un-wA_1OH-tQIUIEHEDrnS_-}{https://drive.google.com/drive/folders/1axgu3-65un-wA\_1OH-tQIUIEHEDrnS\_-}








\bibliography{references}


\begin{thebibliography}{54}
\ifx \bisbn   \undefined \def \bisbn  #1{ISBN #1}\fi
\ifx \binits  \undefined \def \binits#1{#1}\fi
\ifx \bauthor  \undefined \def \bauthor#1{#1}\fi
\ifx \batitle  \undefined \def \batitle#1{#1}\fi
\ifx \bjtitle  \undefined \def \bjtitle#1{#1}\fi
\ifx \bvolume  \undefined \def \bvolume#1{\textbf{#1}}\fi
\ifx \byear  \undefined \def \byear#1{#1}\fi
\ifx \bissue  \undefined \def \bissue#1{#1}\fi
\ifx \bfpage  \undefined \def \bfpage#1{#1}\fi
\ifx \blpage  \undefined \def \blpage #1{#1}\fi
\ifx \burl  \undefined \def \burl#1{\textsf{#1}}\fi
\ifx \doiurl  \undefined \def \doiurl#1{\url{https://doi.org/#1}}\fi
\ifx \betal  \undefined \def \betal{\textit{et al.}}\fi
\ifx \binstitute  \undefined \def \binstitute#1{#1}\fi
\ifx \binstitutionaled  \undefined \def \binstitutionaled#1{#1}\fi
\ifx \bctitle  \undefined \def \bctitle#1{#1}\fi
\ifx \beditor  \undefined \def \beditor#1{#1}\fi
\ifx \bpublisher  \undefined \def \bpublisher#1{#1}\fi
\ifx \bbtitle  \undefined \def \bbtitle#1{#1}\fi
\ifx \bedition  \undefined \def \bedition#1{#1}\fi
\ifx \bseriesno  \undefined \def \bseriesno#1{#1}\fi
\ifx \blocation  \undefined \def \blocation#1{#1}\fi
\ifx \bsertitle  \undefined \def \bsertitle#1{#1}\fi
\ifx \bsnm \undefined \def \bsnm#1{#1}\fi
\ifx \bsuffix \undefined \def \bsuffix#1{#1}\fi
\ifx \bparticle \undefined \def \bparticle#1{#1}\fi
\ifx \barticle \undefined \def \barticle#1{#1}\fi
\bibcommenthead
\ifx \bconfdate \undefined \def \bconfdate #1{#1}\fi
\ifx \botherref \undefined \def \botherref #1{#1}\fi
\ifx \url \undefined \def \url#1{\textsf{#1}}\fi
\ifx \bchapter \undefined \def \bchapter#1{#1}\fi
\ifx \bbook \undefined \def \bbook#1{#1}\fi
\ifx \bcomment \undefined \def \bcomment#1{#1}\fi
\ifx \oauthor \undefined \def \oauthor#1{#1}\fi
\ifx \citeauthoryear \undefined \def \citeauthoryear#1{#1}\fi
\ifx \endbibitem  \undefined \def \endbibitem {}\fi
\ifx \bconflocation  \undefined \def \bconflocation#1{#1}\fi
\ifx \arxivurl  \undefined \def \arxivurl#1{\textsf{#1}}\fi
\csname PreBibitemsHook\endcsname

\bibitem[\protect\citeauthoryear{Hao et~al.}{2020}]{seg_survey1}
\begin{barticle}
\bauthor{\bsnm{Hao}, \binits{S.}},
\bauthor{\bsnm{Zhou}, \binits{Y.}},
\bauthor{\bsnm{Guo}, \binits{Y.}}:
\batitle{A brief survey on semantic segmentation with deep learning}.
\bjtitle{Neurocomputing}
\bvolume{406},
\bfpage{302}--\blpage{321}
(\byear{2020})
\end{barticle}
\endbibitem

\bibitem[\protect\citeauthoryear{Sankaranarayanan et~al.}{2018}]{domain_shift_1}
\begin{bchapter}
\bauthor{\bsnm{Sankaranarayanan}, \binits{S.}},
\bauthor{\bsnm{Balaji}, \binits{Y.}},
\bauthor{\bsnm{Jain}, \binits{A.}},
\bauthor{\bsnm{Lim}, \binits{S.N.}},
\bauthor{\bsnm{Chellappa}, \binits{R.}}:
\bctitle{Learning from synthetic data: Addressing domain shift for semantic segmentation}.
In: \bbtitle{IEEE Conference on Computer Vision and Pattern Recognition},
pp. \bfpage{3752}--\blpage{3761}
(\byear{2018})
\end{bchapter}
\endbibitem

\bibitem[\protect\citeauthoryear{Zhang et~al.}{2022}]{Parauda}
\begin{barticle}
\bauthor{\bsnm{Zhang}, \binits{W.}},
\bauthor{\bsnm{Wang}, \binits{J.}},
\bauthor{\bsnm{Wang}, \binits{Y.}},
\bauthor{\bsnm{Wang}, \binits{F.-Y.}}:
\batitle{{Parauda}: Invariant feature learning with auxiliary synthetic samples for unsupervised domain adaptation}.
\bjtitle{IEEE Transactions on Intelligent Transportation Systems}
\bvolume{23}(\bissue{11}),
\bfpage{20217}--\blpage{20229}
(\byear{2022})
\end{barticle}
\endbibitem

\bibitem[\protect\citeauthoryear{Sun et~al.}{2022}]{ODADA}
\begin{barticle}
\bauthor{\bsnm{Sun}, \binits{Y.}},
\bauthor{\bsnm{Dai}, \binits{D.}},
\bauthor{\bsnm{Xu}, \binits{S.}}:
\batitle{Rethinking adversarial domain adaptation: Orthogonal decomposition for unsupervised domain adaptation in medical image segmentation}.
\bjtitle{Medical Image Analysis}
\bvolume{82},
\bfpage{102623}
(\byear{2022})
\end{barticle}
\endbibitem

\bibitem[\protect\citeauthoryear{Hu et~al.}{2022}]{DoCR}
\begin{bchapter}
\bauthor{\bsnm{Hu}, \binits{S.}},
\bauthor{\bsnm{Liao}, \binits{Z.}},
\bauthor{\bsnm{Xia}, \binits{Y.}}:
\bctitle{Domain specific convolution and high frequency reconstruction based unsupervised domain adaptation for medical image segmentation}.
In: \bbtitle{International Conference on Medical Image Computing and Computer-Assisted Intervention},
pp. \bfpage{650}--\blpage{659}
(\byear{2022}).
\bcomment{Springer}
\end{bchapter}
\endbibitem

\bibitem[\protect\citeauthoryear{Shin et~al.}{2021}]{UDA_Disentangle}
\begin{bchapter}
\bauthor{\bsnm{Shin}, \binits{S.Y.}},
\bauthor{\bsnm{Lee}, \binits{S.}},
\bauthor{\bsnm{Summers}, \binits{R.M.}}:
\bctitle{Unsupervised domain adaptation for small bowel segmentation using disentangled representation}.
In: \bbtitle{International Conference on Medical Image Computing and Computer-Assisted Intervention},
pp. \bfpage{282}--\blpage{292}
(\byear{2021}).
\bcomment{Springer}
\end{bchapter}
\endbibitem

\bibitem[\protect\citeauthoryear{Wang et~al.}{2019}]{BEAL}
\begin{bchapter}
\bauthor{\bsnm{Wang}, \binits{S.}},
\bauthor{\bsnm{Yu}, \binits{L.}},
\bauthor{\bsnm{Li}, \binits{K.}},
\bauthor{\bsnm{Yang}, \binits{X.}},
\bauthor{\bsnm{Fu}, \binits{C.-W.}},
\bauthor{\bsnm{Heng}, \binits{P.-A.}}:
\bctitle{Boundary and entropy-driven adversarial learning for fundus image segmentation}.
In: \bbtitle{International Conference on Medical Image Computing and Computer-Assisted Intervention},
pp. \bfpage{102}--\blpage{110}
(\byear{2019}).
\bcomment{Springer}
\end{bchapter}
\endbibitem

\bibitem[\protect\citeauthoryear{Chen et~al.}{2022}]{DALN}
\begin{bchapter}
\bauthor{\bsnm{Chen}, \binits{L.}},
\bauthor{\bsnm{Chen}, \binits{H.}},
\bauthor{\bsnm{Wei}, \binits{Z.}},
\bauthor{\bsnm{Jin}, \binits{X.}},
\bauthor{\bsnm{Tan}, \binits{X.}},
\bauthor{\bsnm{Jin}, \binits{Y.}},
\bauthor{\bsnm{Chen}, \binits{E.}}:
\bctitle{Reusing the task-specific classifier as a discriminator: Discriminator-free adversarial domain adaptation}.
In: \bbtitle{IEEE/CVF Conference on Computer Vision and Pattern Recognition},
pp. \bfpage{7181}--\blpage{7190}
(\byear{2022})
\end{bchapter}
\endbibitem

\bibitem[\protect\citeauthoryear{Li et~al.}{2024}]{SFDA_survey1}
\begin{botherref}
\oauthor{\bsnm{Li}, \binits{J.}},
\oauthor{\bsnm{Yu}, \binits{Z.}},
\oauthor{\bsnm{Du}, \binits{Z.}},
\oauthor{\bsnm{Zhu}, \binits{L.}},
\oauthor{\bsnm{Shen}, \binits{H.T.}}:
A comprehensive survey on source-free domain adaptation.
IEEE Transactions on Pattern Analysis and Machine Intelligence
(2024)
\end{botherref}
\endbibitem

\bibitem[\protect\citeauthoryear{Liang et~al.}{2023}]{TTA_survey}
\begin{botherref}
\oauthor{\bsnm{Liang}, \binits{J.}},
\oauthor{\bsnm{He}, \binits{R.}},
\oauthor{\bsnm{Tan}, \binits{T.}}:
A comprehensive survey on test-time adaptation under distribution shifts.
arXiv preprint arXiv:2303.15361
(2023)
\end{botherref}
\endbibitem

\bibitem[\protect\citeauthoryear{Wang et~al.}{2021}]{TENT}
\begin{bchapter}
\bauthor{\bsnm{Wang}, \binits{D.}},
\bauthor{\bsnm{Shelhamer}, \binits{E.}},
\bauthor{\bsnm{Liu}, \binits{S.}},
\bauthor{\bsnm{Olshausen}, \binits{B.}},
\bauthor{\bsnm{Darrell}, \binits{T.}}:
\bctitle{tent: fully test-time adaptation by entropy minimization}.
In: \bbtitle{International Conference on Learning Representations}
(\byear{2021})
\end{bchapter}
\endbibitem

\bibitem[\protect\citeauthoryear{Chaurasia et~al.}{2022}]{storedata1}
\begin{botherref}
\oauthor{\bsnm{Chaurasia}, \binits{M.A.}},
\oauthor{\bsnm{Moin}, \binits{M.U.}},
\oauthor{\bsnm{Uddin}, \binits{S.A.}},
\oauthor{\bsnm{Shoaib}, \binits{M.A.}}:
Personal cloud system for hospital data management to store covid-19 patients records.
Contactless Healthcare Facilitation and Commodity Delivery Management During COVID 19 Pandemic,
49--77
(2022)
\end{botherref}
\endbibitem

\bibitem[\protect\citeauthoryear{Li et~al.}{2016}]{storedata2}
\begin{botherref}
\oauthor{\bsnm{Li}, \binits{J.-S.}},
\oauthor{\bsnm{Zhang}, \binits{Y.-F.}},
\oauthor{\bsnm{Tian}, \binits{Y.}}:
Medical big data analysis in hospital information system.
Big data on real-world applications
\textbf{65}
(2016)
\end{botherref}
\endbibitem

\bibitem[\protect\citeauthoryear{Chen et~al.}{2024}]{VPTTA}
\begin{bchapter}
\bauthor{\bsnm{Chen}, \binits{Z.}},
\bauthor{\bsnm{Pan}, \binits{Y.}},
\bauthor{\bsnm{Ye}, \binits{Y.}},
\bauthor{\bsnm{Lu}, \binits{M.}},
\bauthor{\bsnm{Xia}, \binits{Y.}}:
\bctitle{Each test image deserves a specific prompt: Continual test-time adaptation for 2d medical image segmentation}.
In: \bbtitle{IEEE/CVF Conference on Computer Vision and Pattern Recognition},
pp. \bfpage{11184}--\blpage{11193}
(\byear{2024})
\end{bchapter}
\endbibitem

\bibitem[\protect\citeauthoryear{Zhu et~al.}{2017}]{cyclegan}
\begin{bchapter}
\bauthor{\bsnm{Zhu}, \binits{J.-Y.}},
\bauthor{\bsnm{Park}, \binits{T.}},
\bauthor{\bsnm{Isola}, \binits{P.}},
\bauthor{\bsnm{Efros}, \binits{A.A.}}:
\bctitle{Unpaired image-to-image translation using cycle-consistent adversarial networks}.
In: \bbtitle{IEEE International Conference on Computer Vision},
pp. \bfpage{2223}--\blpage{2232}
(\byear{2017})
\end{bchapter}
\endbibitem

\bibitem[\protect\citeauthoryear{Wang and Zheng}{2022}]{CYCMIS}
\begin{barticle}
\bauthor{\bsnm{Wang}, \binits{R.}},
\bauthor{\bsnm{Zheng}, \binits{G.}}:
\batitle{{CyCMIS}: Cycle-consistent cross-domain medical image segmentation via diverse image augmentation}.
\bjtitle{Medical Image Analysis}
\bvolume{76},
\bfpage{102328}
(\byear{2022})
\end{barticle}
\endbibitem

\bibitem[\protect\citeauthoryear{Sanchez et~al.}{2022}]{CXDaGAN}
\begin{botherref}
\oauthor{\bsnm{Sanchez}, \binits{K.}},
\oauthor{\bsnm{Hinojosa}, \binits{C.}},
\oauthor{\bsnm{Arguello}, \binits{H.}},
\oauthor{\bsnm{Kouam{\'e}}, \binits{D.}},
\oauthor{\bsnm{Meyrignac}, \binits{O.}},
\oauthor{\bsnm{Basarab}, \binits{A.}}:
{CX-DaGAN}: Domain adaptation for pneumonia diagnosis on a small chest x-ray dataset.
IEEE Transactions on Medical Imaging
(2022)
\end{botherref}
\endbibitem

\bibitem[\protect\citeauthoryear{Yang and Soatto}{2020}]{FDA}
\begin{bchapter}
\bauthor{\bsnm{Yang}, \binits{Y.}},
\bauthor{\bsnm{Soatto}, \binits{S.}}:
\bctitle{{FDA}: Fourier domain adaptation for semantic segmentation}.
In: \bbtitle{IEEE/CVF Conference on Computer Vision and Pattern Recognition},
pp. \bfpage{4085}--\blpage{4095}
(\byear{2020})
\end{bchapter}
\endbibitem

\bibitem[\protect\citeauthoryear{Kumar et~al.}{2021}]{CAFT}
\begin{bchapter}
\bauthor{\bsnm{Kumar}, \binits{V.}},
\bauthor{\bsnm{Srivastava}, \binits{S.}},
\bauthor{\bsnm{Lal}, \binits{R.}},
\bauthor{\bsnm{Chakraborty}, \binits{A.}}:
\bctitle{{CAFT}: Class aware frequency transform for reducing domain gap}.
In: \bbtitle{IEEE/CVF International Conference on Computer Vision},
pp. \bfpage{2525}--\blpage{2534}
(\byear{2021})
\end{bchapter}
\endbibitem

\bibitem[\protect\citeauthoryear{Wang et~al.}{2021}]{gaussian_fa}
\begin{barticle}
\bauthor{\bsnm{Wang}, \binits{J.}},
\bauthor{\bsnm{Chen}, \binits{J.}},
\bauthor{\bsnm{Lin}, \binits{J.}},
\bauthor{\bsnm{Sigal}, \binits{L.}},
\bauthor{\bsnm{Silva}, \binits{C.W.}}:
\batitle{Discriminative feature alignment: Improving transferability of unsupervised domain adaptation by gaussian-guided latent alignment}.
\bjtitle{Pattern Recognition}
\bvolume{116},
\bfpage{107943}
(\byear{2021})
\end{barticle}
\endbibitem

\bibitem[\protect\citeauthoryear{Liu et~al.}{2022}]{ECSD}
\begin{barticle}
\bauthor{\bsnm{Liu}, \binits{B.}},
\bauthor{\bsnm{Pan}, \binits{D.}},
\bauthor{\bsnm{Shuai}, \binits{Z.}},
\bauthor{\bsnm{Song}, \binits{H.}}:
\batitle{{ECSD-Net}: A joint optic disc and cup segmentation and glaucoma classification network based on unsupervised domain adaptation}.
\bjtitle{Computer Methods and Programs in Biomedicine}
\bvolume{213},
\bfpage{106530}
(\byear{2022})
\end{barticle}
\endbibitem

\bibitem[\protect\citeauthoryear{Sun et~al.}{2020}]{TTT}
\begin{bchapter}
\bauthor{\bsnm{Sun}, \binits{Y.}},
\bauthor{\bsnm{Wang}, \binits{X.}},
\bauthor{\bsnm{Liu}, \binits{Z.}},
\bauthor{\bsnm{Miller}, \binits{J.}},
\bauthor{\bsnm{Efros}, \binits{A.}},
\bauthor{\bsnm{Hardt}, \binits{M.}}:
\bctitle{Test-time training with self-supervision for generalization under distribution shifts}.
In: \bbtitle{International Conference on Machine Learning},
pp. \bfpage{9229}--\blpage{9248}
(\byear{2020}).
\bcomment{PMLR}
\end{bchapter}
\endbibitem

\bibitem[\protect\citeauthoryear{Niu et~al.}{2023}]{SAR}
\begin{bchapter}
\bauthor{\bsnm{Niu}, \binits{S.}},
\bauthor{\bsnm{Wu}, \binits{J.}},
\bauthor{\bsnm{Zhang}, \binits{Y.}},
\bauthor{\bsnm{Wen}, \binits{Z.}},
\bauthor{\bsnm{Chen}, \binits{Y.}},
\bauthor{\bsnm{Zhao}, \binits{P.}},
\bauthor{\bsnm{Tan}, \binits{M.}}:
\bctitle{Towards stable test-time adaptation in dynamic wild world}.
In: \bbtitle{The Eleventh International Conference on Learning Representations}
(\byear{2023})
\end{bchapter}
\endbibitem

\bibitem[\protect\citeauthoryear{Yang et~al.}{2022}]{DLTTA}
\begin{barticle}
\bauthor{\bsnm{Yang}, \binits{H.}},
\bauthor{\bsnm{Chen}, \binits{C.}},
\bauthor{\bsnm{Jiang}, \binits{M.}},
\bauthor{\bsnm{Liu}, \binits{Q.}},
\bauthor{\bsnm{Cao}, \binits{J.}},
\bauthor{\bsnm{Heng}, \binits{P.A.}},
\bauthor{\bsnm{Dou}, \binits{Q.}}:
\batitle{Dltta: Dynamic learning rate for test-time adaptation on cross-domain medical images}.
\bjtitle{IEEE Transactions on Image Processing}
\bvolume{41}(\bissue{12}),
\bfpage{3575}--\blpage{3586}
(\byear{2022})
\end{barticle}
\endbibitem

\bibitem[\protect\citeauthoryear{Zhang et~al.}{2023}]{DomainAdaptor}
\begin{bchapter}
\bauthor{\bsnm{Zhang}, \binits{J.}},
\bauthor{\bsnm{Qi}, \binits{L.}},
\bauthor{\bsnm{Shi}, \binits{Y.}},
\bauthor{\bsnm{Gao}, \binits{Y.}}:
\bctitle{Domainadaptor: A novel approach to test-time adaptation}.
In: \bbtitle{IEEE/CVF Conference on Computer Vision and Pattern Recognition},
pp. \bfpage{18971}--\blpage{18981}
(\byear{2023})
\end{bchapter}
\endbibitem

\bibitem[\protect\citeauthoryear{Wang et~al.}{2023}]{Dynamically}
\begin{bchapter}
\bauthor{\bsnm{Wang}, \binits{W.}},
\bauthor{\bsnm{Zhong}, \binits{Z.}},
\bauthor{\bsnm{Wang}, \binits{W.}},
\bauthor{\bsnm{Chen}, \binits{X.}},
\bauthor{\bsnm{Ling}, \binits{C.}},
\bauthor{\bsnm{Wang}, \binits{B.}},
\bauthor{\bsnm{Sebe}, \binits{N.}}:
\bctitle{Dynamically instance-guided adaptation: A backward-free approach for test-time domain adaptive semantic segmentation}.
In: \bbtitle{IEEE/CVF Conference on Computer Vision and Pattern Recognition},
pp. \bfpage{24090}--\blpage{24099}
(\byear{2023})
\end{bchapter}
\endbibitem

\bibitem[\protect\citeauthoryear{Nado et~al.}{2020}]{BN}
\begin{botherref}
\oauthor{\bsnm{Nado}, \binits{Z.}},
\oauthor{\bsnm{Padhy}, \binits{S.}},
\oauthor{\bsnm{Sculley}, \binits{D.}},
\oauthor{\bsnm{D'Amour}, \binits{A.}},
\oauthor{\bsnm{Lakshminarayanan}, \binits{B.}},
\oauthor{\bsnm{Snoek}, \binits{J.}}:
Evaluating prediction-time batch normalization for robustness under covariate shift.
arXiv preprint arXiv:2006.10963
(2020)
\end{botherref}
\endbibitem

\bibitem[\protect\citeauthoryear{Mirza et~al.}{2022}]{DUA}
\begin{bchapter}
\bauthor{\bsnm{Mirza}, \binits{M.J.}},
\bauthor{\bsnm{Micorek}, \binits{J.}},
\bauthor{\bsnm{Possegger}, \binits{H.}},
\bauthor{\bsnm{Bischof}, \binits{H.}}:
\bctitle{The norm must go on: Dynamic unsupervised domain adaptation by normalization}.
In: \bbtitle{IEEE/CVF Conference on Computer Vision and Pattern Recognition},
pp. \bfpage{14765}--\blpage{14775}
(\byear{2022})
\end{bchapter}
\endbibitem

\bibitem[\protect\citeauthoryear{Tian et~al.}{2021}]{VDM_DA}
\begin{barticle}
\bauthor{\bsnm{Tian}, \binits{J.}},
\bauthor{\bsnm{Zhang}, \binits{J.}},
\bauthor{\bsnm{Li}, \binits{W.}},
\bauthor{\bsnm{Xu}, \binits{D.}}:
\batitle{Vdm-da: Virtual domain modeling for source data-free domain adaptation}.
\bjtitle{IEEE Transactions on Circuits and Systems for Video Technology}
\bvolume{32}(\bissue{6}),
\bfpage{3749}--\blpage{3760}
(\byear{2021})
\end{barticle}
\endbibitem

\bibitem[\protect\citeauthoryear{Yang et~al.}{2022}]{FSM}
\begin{barticle}
\bauthor{\bsnm{Yang}, \binits{C.}},
\bauthor{\bsnm{Guo}, \binits{X.}},
\bauthor{\bsnm{Chen}, \binits{Z.}},
\bauthor{\bsnm{Yuan}, \binits{Y.}}:
\batitle{Source free domain adaptation for medical image segmentation with fourier style mining}.
\bjtitle{Medical Image Analysis}
\bvolume{79},
\bfpage{102457}
(\byear{2022})
\end{barticle}
\endbibitem

\bibitem[\protect\citeauthoryear{Li et~al.}{2020}]{li2020model}
\begin{bchapter}
\bauthor{\bsnm{Li}, \binits{R.}},
\bauthor{\bsnm{Jiao}, \binits{Q.}},
\bauthor{\bsnm{Cao}, \binits{W.}},
\bauthor{\bsnm{Wong}, \binits{H.-S.}},
\bauthor{\bsnm{Wu}, \binits{S.}}:
\bctitle{Model adaptation: Unsupervised domain adaptation without source data}.
In: \bbtitle{IEEE/CVF Conference on Computer Vision and Pattern Recognition},
pp. \bfpage{9641}--\blpage{9650}
(\byear{2020})
\end{bchapter}
\endbibitem

\bibitem[\protect\citeauthoryear{Xia et~al.}{2021}]{xia2021adaptive}
\begin{bchapter}
\bauthor{\bsnm{Xia}, \binits{H.}},
\bauthor{\bsnm{Zhao}, \binits{H.}},
\bauthor{\bsnm{Ding}, \binits{Z.}}:
\bctitle{Adaptive adversarial network for source-free domain adaptation}.
In: \bbtitle{IEEE/CVF International Conference on Computer Vision},
pp. \bfpage{9010}--\blpage{9019}
(\byear{2021})
\end{bchapter}
\endbibitem

\bibitem[\protect\citeauthoryear{Chu et~al.}{2022}]{chu2022denoised}
\begin{bchapter}
\bauthor{\bsnm{Chu}, \binits{T.}},
\bauthor{\bsnm{Liu}, \binits{Y.}},
\bauthor{\bsnm{Deng}, \binits{J.}},
\bauthor{\bsnm{Li}, \binits{W.}},
\bauthor{\bsnm{Duan}, \binits{L.}}:
\bctitle{Denoised maximum classifier discrepancy for source-free unsupervised domain adaptation}.
In: \bbtitle{AAAI Conference on Artificial Intelligence},
vol. \bseriesno{36},
pp. \bfpage{472}--\blpage{480}
(\byear{2022})
\end{bchapter}
\endbibitem

\bibitem[\protect\citeauthoryear{Saito et~al.}{2019}]{saito2019semi}
\begin{bchapter}
\bauthor{\bsnm{Saito}, \binits{K.}},
\bauthor{\bsnm{Kim}, \binits{D.}},
\bauthor{\bsnm{Sclaroff}, \binits{S.}},
\bauthor{\bsnm{Darrell}, \binits{T.}},
\bauthor{\bsnm{Saenko}, \binits{K.}}:
\bctitle{Semi-supervised domain adaptation via minimax entropy}.
In: \bbtitle{IEEE/CVF International Conference on Computer Vision},
pp. \bfpage{8050}--\blpage{8058}
(\byear{2019})
\end{bchapter}
\endbibitem

\bibitem[\protect\citeauthoryear{Li and Zhang}{2018}]{li2018semi}
\begin{barticle}
\bauthor{\bsnm{Li}, \binits{L.}},
\bauthor{\bsnm{Zhang}, \binits{Z.}}:
\batitle{Semi-supervised domain adaptation by covariance matching}.
\bjtitle{IEEE Transactions on Pattern Analysis and Machine Intelligence}
\bvolume{41}(\bissue{11}),
\bfpage{2724}--\blpage{2739}
(\byear{2018})
\end{barticle}
\endbibitem

\bibitem[\protect\citeauthoryear{Wu et~al.}{2023}]{UPL-SFDA}
\begin{botherref}
\oauthor{\bsnm{Wu}, \binits{J.}},
\oauthor{\bsnm{Wang}, \binits{G.}},
\oauthor{\bsnm{Gu}, \binits{R.}},
\oauthor{\bsnm{Lu}, \binits{T.}},
\oauthor{\bsnm{Chen}, \binits{Y.}},
\oauthor{\bsnm{Zhu}, \binits{W.}},
\oauthor{\bsnm{Vercauteren}, \binits{T.}},
\oauthor{\bsnm{Ourselin}, \binits{S.}},
\oauthor{\bsnm{Zhang}, \binits{S.}}:
Upl-sfda: Uncertainty-aware pseudo label guided source-free domain adaptation for medical image segmentation.
IEEE Transactions on Medical Imaging
(2023)
\end{botherref}
\endbibitem

\bibitem[\protect\citeauthoryear{Li et~al.}{2024}]{PLPB}
\begin{bchapter}
\bauthor{\bsnm{Li}, \binits{L.}},
\bauthor{\bsnm{Zhou}, \binits{Y.}},
\bauthor{\bsnm{Yang}, \binits{G.}}:
\bctitle{Robust source-free domain adaptation for fundus image segmentation}.
In: \bbtitle{IEEE/CVF Winter Conference on Applications of Computer Vision},
pp. \bfpage{7840}--\blpage{7849}
(\byear{2024})
\end{bchapter}
\endbibitem

\bibitem[\protect\citeauthoryear{Zhang et~al.}{2021}]{zhang2021source}
\begin{bchapter}
\bauthor{\bsnm{Zhang}, \binits{D.}},
\bauthor{\bsnm{Ye}, \binits{M.}},
\bauthor{\bsnm{Xiong}, \binits{L.}},
\bauthor{\bsnm{Li}, \binits{S.}},
\bauthor{\bsnm{Li}, \binits{X.}}:
\bctitle{Source-style transferred mean teacher for source-data free object detection}.
In: \bbtitle{3rd ACM International Conference on Multimedia in Asia},
pp. \bfpage{1}--\blpage{8}
(\byear{2021})
\end{bchapter}
\endbibitem

\bibitem[\protect\citeauthoryear{Liu et~al.}{2023}]{PETS}
\begin{bchapter}
\bauthor{\bsnm{Liu}, \binits{Q.}},
\bauthor{\bsnm{Lin}, \binits{L.}},
\bauthor{\bsnm{Shen}, \binits{Z.}},
\bauthor{\bsnm{Yang}, \binits{Z.}}:
\bctitle{Periodically exchange teacher-student for source-free object detection}.
In: \bbtitle{IEEE/CVF International Conference on Computer Vision},
pp. \bfpage{6414}--\blpage{6424}
(\byear{2023})
\end{bchapter}
\endbibitem

\bibitem[\protect\citeauthoryear{Hu et~al.}{2022}]{ProSFDA}
\begin{botherref}
\oauthor{\bsnm{Hu}, \binits{S.}},
\oauthor{\bsnm{Liao}, \binits{Z.}},
\oauthor{\bsnm{Xia}, \binits{Y.}}:
Prosfda: Prompt learning based source-free domain adaptation for medical image segmentation.
arXiv preprint arXiv:2211.11514
(2022)
\end{botherref}
\endbibitem

\bibitem[\protect\citeauthoryear{Frigo and Johnson}{1998}]{FFTW}
\begin{bchapter}
\bauthor{\bsnm{Frigo}, \binits{M.}},
\bauthor{\bsnm{Johnson}, \binits{S.G.}}:
\bctitle{{FFTW}: An adaptive software architecture for the fft}.
In: \bbtitle{1998 IEEE International Conference on Acoustics, Speech and Signal Processing},
vol. \bseriesno{3},
pp. \bfpage{1381}--\blpage{1384}
(\byear{1998}).
\bcomment{IEEE}
\end{bchapter}
\endbibitem

\bibitem[\protect\citeauthoryear{Fumero et~al.}{2011}]{RIM_ONE_r3}
\begin{bchapter}
\bauthor{\bsnm{Fumero}, \binits{F.}},
\bauthor{\bsnm{Alay{\'o}n}, \binits{S.}},
\bauthor{\bsnm{Sanchez}, \binits{J.L.}},
\bauthor{\bsnm{Sigut}, \binits{J.}},
\bauthor{\bsnm{Gonzalez-Hernandez}, \binits{M.}}:
\bctitle{{RIM-ONE}: An open retinal image database for optic nerve evaluation}.
In: \bbtitle{2011 24th International Symposium on Computer-Based Medical Systems},
pp. \bfpage{1}--\blpage{6}
(\byear{2011}).
\bcomment{IEEE}
\end{bchapter}
\endbibitem

\bibitem[\protect\citeauthoryear{Orlando et~al.}{2020}]{REFUGE}
\begin{barticle}
\bauthor{\bsnm{Orlando}, \binits{J.I.}},
\bauthor{\bsnm{Fu}, \binits{H.}},
\bauthor{\bsnm{Breda}, \binits{J.B.}},
\bauthor{\bsnm{Keer}, \binits{K.}},
\bauthor{\bsnm{Bathula}, \binits{D.R.}},
\bauthor{\bsnm{Diaz-Pinto}, \binits{A.}},
\bauthor{\bsnm{Fang}, \binits{R.}},
\bauthor{\bsnm{Heng}, \binits{P.-A.}},
\bauthor{\bsnm{Kim}, \binits{J.}},
\bauthor{\bsnm{Lee}, \binits{J.}}, \betal:
\batitle{{REFUGE Challenge}: A unified framework for evaluating automated methods for glaucoma assessment from fundus photographs}.
\bjtitle{Medical Image Analysis}
\bvolume{59},
\bfpage{101570}
(\byear{2020})
\end{barticle}
\endbibitem

\bibitem[\protect\citeauthoryear{Zhang et~al.}{2010}]{ORIGA}
\begin{bchapter}
\bauthor{\bsnm{Zhang}, \binits{Z.}},
\bauthor{\bsnm{Yin}, \binits{F.S.}},
\bauthor{\bsnm{Liu}, \binits{J.}},
\bauthor{\bsnm{Wong}, \binits{W.K.}},
\bauthor{\bsnm{Tan}, \binits{N.M.}},
\bauthor{\bsnm{Lee}, \binits{B.H.}},
\bauthor{\bsnm{Cheng}, \binits{J.}},
\bauthor{\bsnm{Wong}, \binits{T.Y.}}:
\bctitle{{ORIGA-light}: An online retinal fundus image database for glaucoma analysis and research}.
In: \bbtitle{2010 Annual International Conference of the IEEE Engineering in Medicine and Biology},
pp. \bfpage{3065}--\blpage{3068}
(\byear{2010}).
\bcomment{IEEE}
\end{bchapter}
\endbibitem

\bibitem[\protect\citeauthoryear{Sivaswamy et~al.}{2014}]{Drishti-GS}
\begin{bchapter}
\bauthor{\bsnm{Sivaswamy}, \binits{J.}},
\bauthor{\bsnm{Krishnadas}, \binits{S.}},
\bauthor{\bsnm{Joshi}, \binits{G.D.}},
\bauthor{\bsnm{Jain}, \binits{M.}},
\bauthor{\bsnm{Tabish}, \binits{A.U.S.}}:
\bctitle{{Drishti-GS}: Retinal image dataset for optic nerve head (onh) segmentation}.
In: \bbtitle{2014 IEEE 11th International Symposium on Biomedical Imaging},
pp. \bfpage{53}--\blpage{56}
(\byear{2014}).
\bcomment{IEEE}
\end{bchapter}
\endbibitem

\bibitem[\protect\citeauthoryear{Ngoc~Lan et~al.}{2021}]{BKAI_IGH_NeoPolyp}
\begin{bchapter}
\bauthor{\bsnm{Ngoc~Lan}, \binits{P.}},
\bauthor{\bsnm{An}, \binits{N.S.}},
\bauthor{\bsnm{Hang}, \binits{D.V.}},
\bauthor{\bsnm{Long}, \binits{D.V.}},
\bauthor{\bsnm{Trung}, \binits{T.Q.}},
\bauthor{\bsnm{Thuy}, \binits{N.T.}},
\bauthor{\bsnm{Sang}, \binits{D.V.}}:
\bctitle{Neounet: Towards accurate colon polyp segmentation and neoplasm detection}.
In: \bbtitle{Advances in Visual Computing: 16th International Symposium, ISVC 2021, Virtual Event, October 4-6, 2021, Proceedings, Part II},
pp. \bfpage{15}--\blpage{28}
(\byear{2021}).
\bcomment{Springer}
\end{bchapter}
\endbibitem

\bibitem[\protect\citeauthoryear{Bernal et~al.}{2015}]{CVC_ClinicDB}
\begin{barticle}
\bauthor{\bsnm{Bernal}, \binits{J.}},
\bauthor{\bsnm{S{\'a}nchez}, \binits{F.J.}},
\bauthor{\bsnm{Fern{\'a}ndez-Esparrach}, \binits{G.}},
\bauthor{\bsnm{Gil}, \binits{D.}},
\bauthor{\bsnm{Rodr{\'\i}guez}, \binits{C.}},
\bauthor{\bsnm{Vilari{\~n}o}, \binits{F.}}:
\batitle{Wm-dova maps for accurate polyp highlighting in colonoscopy: Validation vs. saliency maps from physicians}.
\bjtitle{Computerized Medical Imaging and Graphics}
\bvolume{43},
\bfpage{99}--\blpage{111}
(\byear{2015})
\end{barticle}
\endbibitem

\bibitem[\protect\citeauthoryear{Silva et~al.}{2014}]{ETIS_LaribPolypDB}
\begin{barticle}
\bauthor{\bsnm{Silva}, \binits{J.}},
\bauthor{\bsnm{Histace}, \binits{A.}},
\bauthor{\bsnm{Romain}, \binits{O.}},
\bauthor{\bsnm{Dray}, \binits{X.}},
\bauthor{\bsnm{Granado}, \binits{B.}}:
\batitle{Toward embedded detection of polyps in wce images for early diagnosis of colorectal cancer}.
\bjtitle{International Journal of Computer Assisted Radiology and Surgery}
\bvolume{9}(\bissue{2}),
\bfpage{283}--\blpage{293}
(\byear{2014})
\end{barticle}
\endbibitem

\bibitem[\protect\citeauthoryear{Jha et~al.}{2020}]{Kvasir_Seg}
\begin{bchapter}
\bauthor{\bsnm{Jha}, \binits{D.}},
\bauthor{\bsnm{Smedsrud}, \binits{P.H.}},
\bauthor{\bsnm{Riegler}, \binits{M.A.}},
\bauthor{\bsnm{Halvorsen}, \binits{P.}},
\bauthor{\bsnm{Lange}, \binits{T.d.}},
\bauthor{\bsnm{Johansen}, \binits{D.}},
\bauthor{\bsnm{Johansen}, \binits{H.D.}}:
\bctitle{Kvasir-seg: A segmented polyp dataset}.
In: \bbtitle{MultiMedia Modeling: 26th International Conference, MMM 2020, Daejeon, South Korea, January 5--8, 2020, Proceedings, Part II 26},
pp. \bfpage{451}--\blpage{462}
(\byear{2020}).
\bcomment{Springer}
\end{bchapter}
\endbibitem

\bibitem[\protect\citeauthoryear{Fan et~al.}{2020}]{PraNet}
\begin{bchapter}
\bauthor{\bsnm{Fan}, \binits{D.-P.}},
\bauthor{\bsnm{Ji}, \binits{G.-P.}},
\bauthor{\bsnm{Zhou}, \binits{T.}},
\bauthor{\bsnm{Chen}, \binits{G.}},
\bauthor{\bsnm{Fu}, \binits{H.}},
\bauthor{\bsnm{Shen}, \binits{J.}},
\bauthor{\bsnm{Shao}, \binits{L.}}:
\bctitle{Pranet: Parallel reverse attention network for polyp segmentation}.
In: \bbtitle{International Conference on Medical Image Computing and Computer-Assisted Intervention},
pp. \bfpage{263}--\blpage{273}
(\byear{2020}).
\bcomment{Springer}
\end{bchapter}
\endbibitem

\bibitem[\protect\citeauthoryear{Fan et~al.}{2018}]{Metric_E_max}
\begin{bchapter}
\bauthor{\bsnm{Fan}, \binits{D.-P.}},
\bauthor{\bsnm{Gong}, \binits{C.}},
\bauthor{\bsnm{Cao}, \binits{Y.}},
\bauthor{\bsnm{Ren}, \binits{B.}},
\bauthor{\bsnm{Cheng}, \binits{M.-M.}},
\bauthor{\bsnm{Borji}, \binits{A.}}:
\bctitle{Enhanced-alignment measure for binary foreground map evaluation}.
In: \bbtitle{International Joint Conference on Artificial Intelligence},
pp. \bfpage{698}--\blpage{704}
(\byear{2018})
\end{bchapter}
\endbibitem

\bibitem[\protect\citeauthoryear{Fan et~al.}{2017}]{Metric_S_alpha}
\begin{bchapter}
\bauthor{\bsnm{Fan}, \binits{D.-P.}},
\bauthor{\bsnm{Cheng}, \binits{M.-M.}},
\bauthor{\bsnm{Liu}, \binits{Y.}},
\bauthor{\bsnm{Li}, \binits{T.}},
\bauthor{\bsnm{Borji}, \binits{A.}}:
\bctitle{Structure-measure: A new way to evaluate foreground maps}.
In: \bbtitle{IEEE International Conference on Computer Vision},
pp. \bfpage{4548}--\blpage{4557}
(\byear{2017})
\end{bchapter}
\endbibitem

\bibitem[\protect\citeauthoryear{He et~al.}{2016}]{ResNet}
\begin{bchapter}
\bauthor{\bsnm{He}, \binits{K.}},
\bauthor{\bsnm{Zhang}, \binits{X.}},
\bauthor{\bsnm{Ren}, \binits{S.}},
\bauthor{\bsnm{Sun}, \binits{J.}}:
\bctitle{Deep residual learning for image recognition}.
In: \bbtitle{IEEE Conference on Computer Vision and Pattern Recognition},
pp. \bfpage{770}--\blpage{778}
(\byear{2016})
\end{bchapter}
\endbibitem

\bibitem[\protect\citeauthoryear{Gao et~al.}{2019}]{Res2Net}
\begin{barticle}
\bauthor{\bsnm{Gao}, \binits{S.-H.}},
\bauthor{\bsnm{Cheng}, \binits{M.-M.}},
\bauthor{\bsnm{Zhao}, \binits{K.}},
\bauthor{\bsnm{Zhang}, \binits{X.-Y.}},
\bauthor{\bsnm{Yang}, \binits{M.-H.}},
\bauthor{\bsnm{Torr}, \binits{P.}}:
\batitle{Res2net: A new multi-scale backbone architecture}.
\bjtitle{IEEE Transactions on Pattern Analysis and Machine Intelligence}
\bvolume{43}(\bissue{2}),
\bfpage{652}--\blpage{662}
(\byear{2019})
\end{barticle}
\endbibitem

\end{thebibliography}

\end{document}